\documentclass[10pt,twocolumn,letterpaper]{article}

\usepackage{iccv}
\usepackage{times}
\usepackage{epsfig}
\usepackage{graphicx}
\usepackage{amsmath}
\usepackage{amssymb}
\usepackage{lscape}
\usepackage{appendix}

\usepackage{multirow}
\usepackage{comment}
\usepackage{mathtools}
\usepackage{cuted}
\usepackage[export]{adjustbox}

\newcommand{\twofigures}[3]{
            \centerline{{\includegraphics[width=#3]{#1}}~~{\includegraphics[width=#3]{#2}}}}

\newcommand{\threefigures}[6]{
            \centerline{{\includegraphics[width=#4]{#1}}~~~~{\includegraphics[width=#5]{#2}}~~~~{\includegraphics[width=#6]{#3}}} 
            \makebox[0.82\columnwidth][c]{(a)}\makebox[0.55\columnwidth][c]{(b)}\makebox[0.70\columnwidth][c]{(c)}}

\usepackage{xcolor}
\PassOptionsToPackage{hyphens}{url}\usepackage[pagebackref=true,breaklinks=true,letterpaper=true,colorlinks,
  citecolor=citecolor,bookmarks=false]{hyperref}
\definecolor{citecolor}{RGB}{34,139,34}

\iccvfinalcopy % *** Uncomment this line for the final submission

\def\iccvPaperID{****} % *** Enter the ICCV Paper ID here

\ificcvfinal\fi

\begin{document}

%%%%%%%%% TITLE
\title{Learning Compositional Representations for Few-Shot Recognition}

\author{Pavel Tokmakov\qquad Yu-Xiong Wang\qquad Martial Hebert\\
Robotics Institute, Carnegie Mellon University\\
{\tt\small \{ptokmako,yuxiongw,hebert\}@cs.cmu.edu}
}
%{\small {\url{https://sites.google.com/view/comprepr/home}}}

\maketitle
\ificcvfinal\thispagestyle{empty}\fi

%%%%%%%%% ABSTRACT
\begin{abstract}
One of the key limitations of modern deep learning approaches lies in the amount of data required to train them. Humans, by contrast, can learn to recognize novel categories from just a few examples. Instrumental to this rapid learning ability is the compositional structure of concept representations in the human brain --- something that deep learning models are lacking. In this work, we make a step towards bridging this gap between human and machine learning by introducing a simple regularization technique that allows the learned representation to be decomposable into parts. Our method uses category-level attribute annotations to disentangle the feature space of a network into subspaces corresponding to the attributes. These attributes can be either purely visual, like object parts, or more abstract, like openness and symmetry. We demonstrate the value of compositional representations on three datasets: CUB-200-2011, SUN397, and ImageNet, and show that they require fewer examples to learn classifiers for novel categories. Our code and trained models together with the collected attribute annotations are available at {\url{https://sites.google.com/view/comprepr/home}}.%on the other hand as well as
\end{abstract}

%%%%%%%%% BODY TEXT
\section{Introduction}
\label{sec:intro}
\begin{figure}[t]
\begin{center}
\twofigures{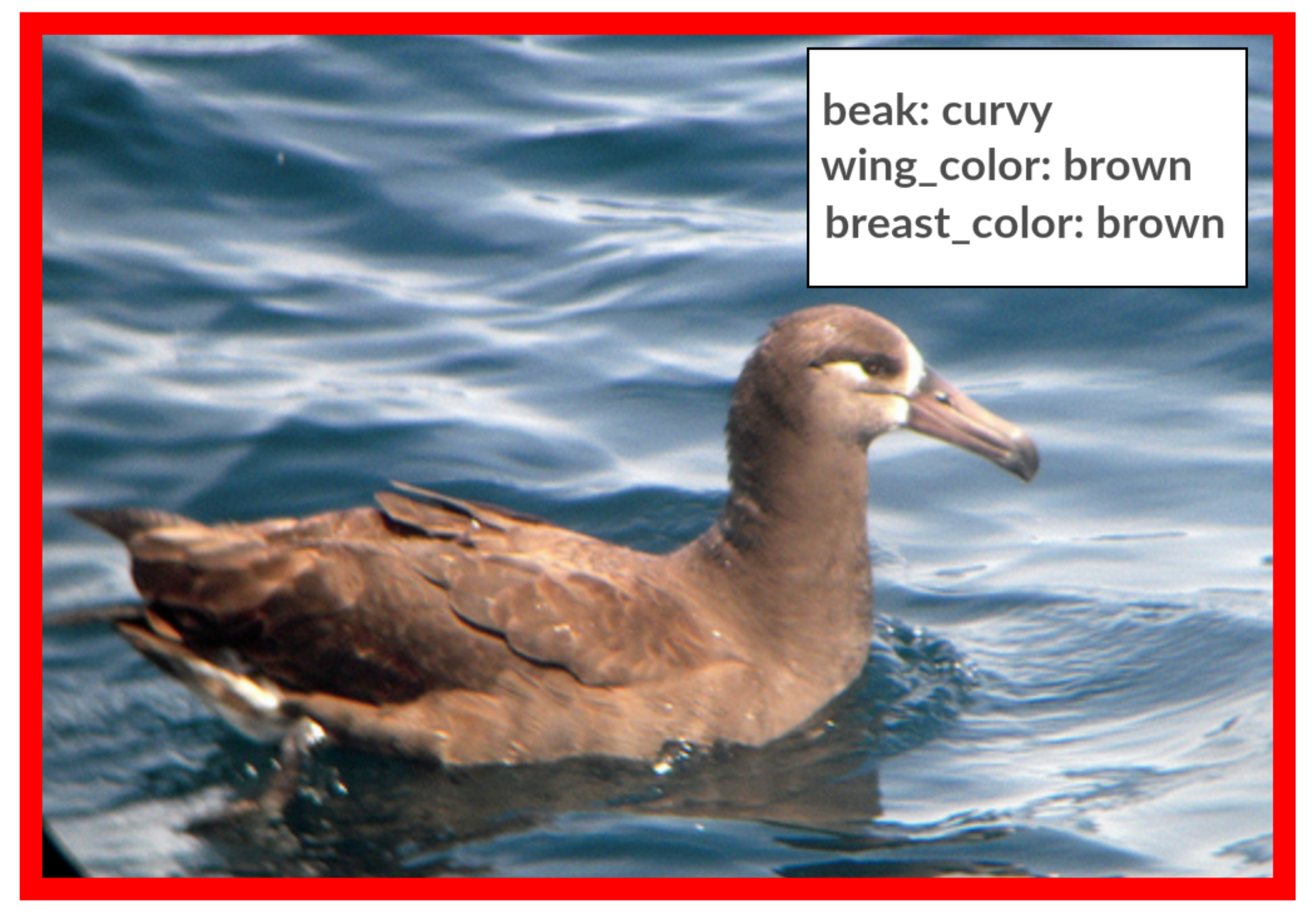}{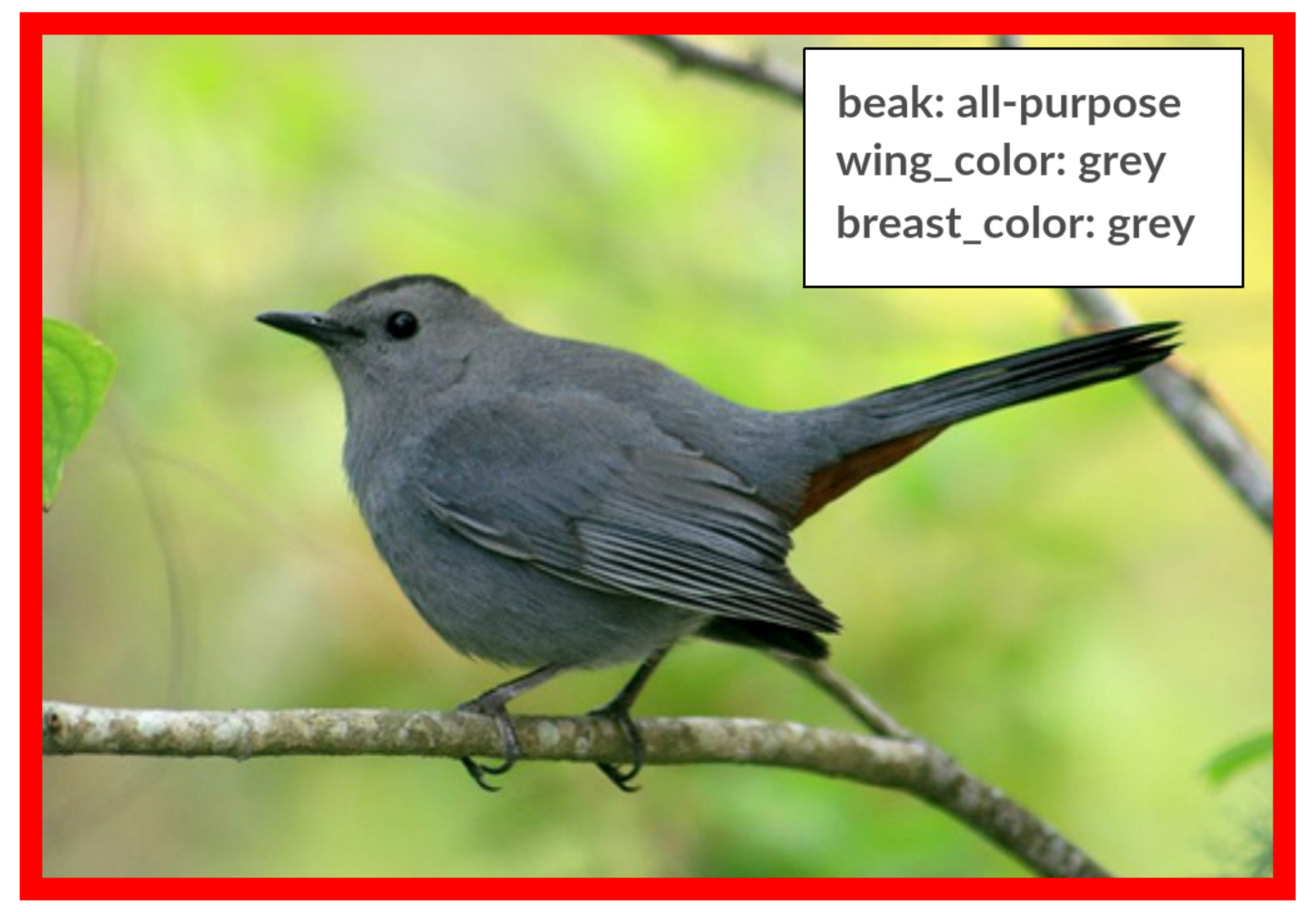}{0.45\columnwidth} %\vspace{0.1cm}
\twofigures{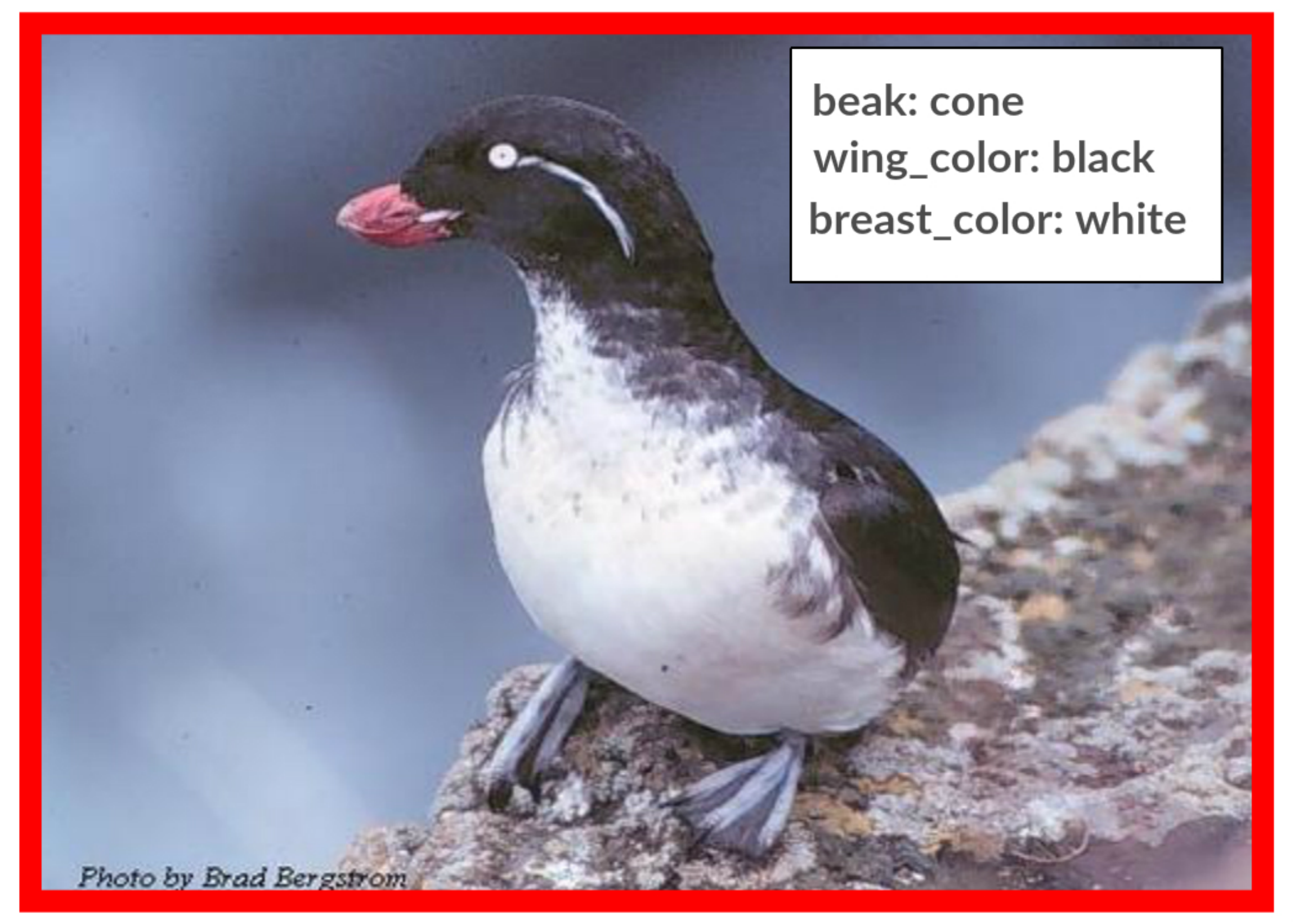}{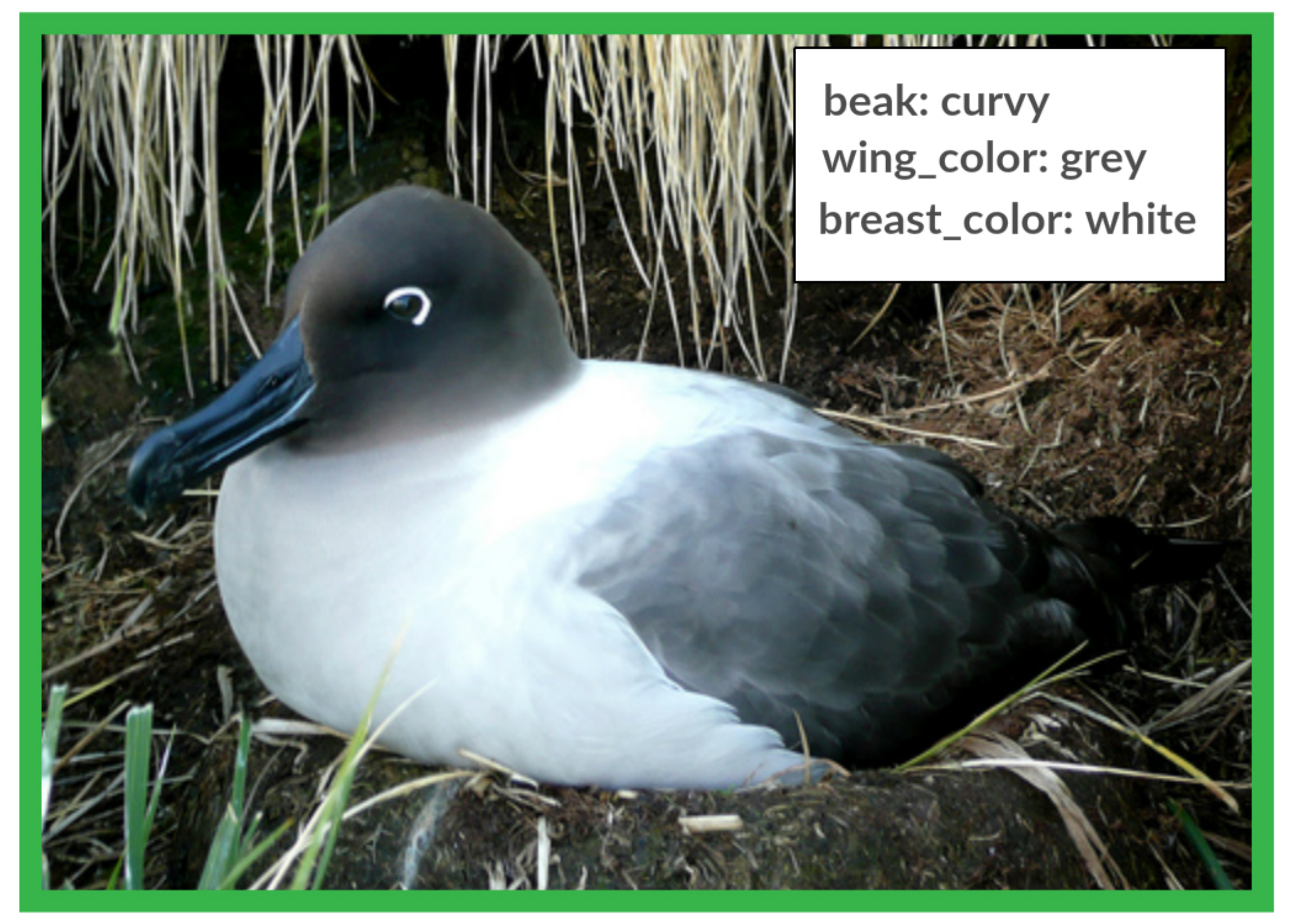}{0.45\columnwidth} \vspace{-.8cm}
\end{center}
\caption{Images from four categories of the CUB-200-2011 dataset, together with some of their attribute annotations. We propose to learn image representations that are decomposable over the attributes. These representations can thus be used to recognize new categories from few examples.}
\label{fig:intro}
\vspace{-0.5cm}
\end{figure}

Consider the images representing four categories from the CUB-200-2011 dataset~\cite{wah2011caltech} in Figure~\ref{fig:intro}. Given a representation learned using the first three categories, shown in red, can a classifier for the fourth category, shown in green, be learned from just a few, or even a single example? This is a problem known as few-shot learning~\cite{thrun1996learning,lake2015human,koch2015siamese,wang2016learningto,finn2017model}. Clearly, it depends on the properties of the representation.
Cognitive science identifies {\em compositionality} as a property that is crucial to this task. Human representations of concepts are decomposable into parts~\cite{biederman1987recognition,hoffman1984parts}, such as the ones shown in the top right corners of the images in Figure~\ref{fig:intro}, allowing classifiers to be rapidly learned for novel concepts through combinations of known primitives~\cite{fodor1975language}. Taking the novel bird category as an example, all of its discriminative attributes have already been observed in the first three categories. These ideas have been highly influential in computer vision, with some of the first models for visual concepts being built as compositions of parts and relations~\cite{marr1982vision,marr1978representation,winston1970learning}.%literature (see the example of the novel bird category --- all of its discriminative attributes have already been observed in the first three categories).

However, state-of-the-art methods for virtually all visual recognition tasks are based on deep learning~\cite{lecun1995convolutional,krizhevsky2012imagenet}. The parameters of deep neural networks are optimized for the end task with gradient-based methods, resulting in representations that are not easily interpretable. There has been a lot of effort on qualitative interpretation of these representations~\cite{zeiler2014visualizing,zhou2018interpreting}, demonstrating that some of the neurons represent object parts. 
Very recently, a quantitative approach to evaluating the compositionality of deep representations has been proposed~\cite{x2018comp}. 
%The authors posit that a feature encoding of an image is compositional over a set of attributes if it can be represented as a sum of the encodings of the attributes.
Nevertheless, these approaches do not investigate the problem of improving the compositional properties of neural networks. In this paper, we propose a simple regularization technique that forces deep image representations to be decomposable into parts, and we empirically demonstrate that such representations facilitate learning classifiers for novel concepts from fewer examples.%, however,

Our method takes as input a dataset of images together with their class labels and {\em category-level} attribute annotations. The attributes can be either purely visual, such as object parts ({\tt beak\_shape}) and scene elements ({\tt grass}), or more abstract, such as {\tt openness} of a scene. 
In~\cite{x2018comp} a feature encoding of an image is defined as compositional over a set of attributes if it can be represented as a combination of the encodings of these attributes. Following this definition, we propose to use attribute annotations as constraints when learning the image representation. This results in a method that, given an image with its corresponding attribute annotations, jointly learns a convolutional neural network (CNN) for the image embedding and a linear layer for the attribute embedding. The attribute embeddings are then used to constrain the image representation to be equal to the sum of the attribute representations (see Figure~\ref{fig:method}(b)).%Taken literally, 

This constraint, however, implies that exhaustive attribute annotations are available. Such an assumption is not realistic for most of the image domains. To address this issue, we propose a relaxed version of the compositionality regularizer. Instead of requiring the image representation to be exactly equal to the sum of the attribute embeddings, it simply maximizes the sum of the individual similarities between the attribute and image embeddings (see Figure~\ref{fig:method}(c)). This ensures that the image representation reflects the compositional structure of the categories, while allowing it to model the remaining factors of variation which are not captured in the annotations. Finally, we observe that enforcing orthogonality of the attribute embeddings leads to a better disentanglement of the resulting image representation.
 
%Given a $d$-dimensional deep image embedding, for it to be fully decomposable over a set of attributes, each dimension has to correspond to one of the attributes (see Figure~\ref{fig:method} (a)). Notice, however, that this definition is based on assumptions that the vocabulary of attributes is exhaustive, and that each attribute can be modeled by a single network unit. In practice, collecting exhaustive attribute annotations is unrealistic for any non-trivial domain, and groups of neurons can better capture fine-grained concepts. Thus, we propose to assign $k$ dimensions of a $n$-dimensional image embedding to each of the attributes, where $k \ll n$, leaving some "free" neurons to model the remaining factors of variation in the data (see Figure~\ref{fig:method} (b)). To enforce these constraints we introduce a simple regularization term in a loss function. Intuitively, applied together with a classification loss, this regularization forces the optimization to chose a representation that is more compositional over the attributes out of the space of all possible discriminative representations.

We evaluate our compositional representation in a few-shot setting on three datasets of different sizes and domains: CUB-200-2011~\cite{wah2011caltech} for fine-grained recognition, SUN397 for scene classification~\cite{xiao2010sun}, and ImageNet~\cite{deng2009imagenet} for object classification. 
%We demonstrate that the proposed regularization \textit{does not simply improve the overall performance of the model}. 
When many training examples are available, it performs on par with the baseline which trains a plain classifier without attribute supervision, but \textit{in the few-shot setting it shows a much better generalization behavior}. In particular, our model achieves an 8\% top-5 accuracy improvement over the baseline in the most challenging 1-shot scenario on SUN397.%learned with the proposed regularization 
%Overall, we demonstrate state-of-the-art results on all three datasets in a variety of settings. our model achieves an 8\% top-5 accuracy improvement over the baseline

An obvious limitation of our approach is that it requires additional annotations. One might ask, how expensive it is to collect the attribute labels, and more importantly, how to even define the vocabulary of attributes for an arbitrary dataset. 
To illustrate that collecting category-level attributes is in fact relatively easy even for large-scale datasets, we label 159 attributes for a subset of the ImageNet categories defined in~\cite{hariharan2017low,wang2018low}. A crucial detail is that the attributes have to be labeled on the category, not on the image level, which allowed us to collect the annotations in just {\em three} days. In addition, note that our approach {\em does not require attribute annotations for novel classes}.% Our collected attribute annotations together with the code and trained models are available on the project page\footnote{\url{https://sites.google.com/view/comprepr/home}}.% More details on the annotations process are provided in Section~\ref{sec:imnet_attrs}.

{\bf Our contributions} are three-fold. (1) We propose the first approach for learning deep compositional representations in Section~\ref{sec:method}. Our method takes images together with their attribute annotations as input and applies a regularizer to enforce the image representation to be decomposable over the attributes. (2) We illustrate the simplicity of collecting attribute annotations on a subset of the ImageNet dataset in Section~\ref{sec:imnet_attrs}. (3) We provide a comprehensive analysis of the learned representation in the context of few-shot learning on three datasets. The evaluation in Section~\ref{sec:exp} demonstrates that our proposed approach results in a representation that generalizes significantly better and \textit{requires fewer examples to learn novel categories}.
%\vspace{-0.2cm}
\section{Related Work}
\label{sec:related}
{\bf Few-shot learning} is a classic problem of recognition with only a few training examples~\cite{thrun1996learning}. Lake \etal~\cite{lake2015human} explicitly encode compositionality and causality properties with Bayesian probabilistic programs. Learning then boils down to constructing programs that best explain the observations and can be done efficiently with a single example per category. However, this approach is limited by that the programs have to be manually defined for each new domain.%proposed to  the fact 

State-of-the-art methods for few-shot learning can be categorized into the ones based on metric learning~\cite{koch2015siamese,vinyals2016matching,snell2017prototypical,yang2018learning} --- training a network to predict whether two images belong to the same category, and the ones built around the idea of meta-learning~\cite{finn2017model,ravi2016optimization,wang2016learningfrom,wang2016learningto} --- training with a loss that explicitly enforces easy adaptation of the weights to new categories with only a few examples. Separately from these approaches, some work proposes to learn to generate additional examples for unseen categories~\cite{wang2018low,hariharan2017low}. Recently, it has been shown that it is crucial to use cosine similarity as a distance measure to achieve top results in few-shot learning evaluation~\cite{gidaris2018dynamic}. Even more recently, Chen \etal~\cite{x2018fewshot} demonstrate that a simple baseline approach --- a linear layer learned on top of a frozen CNN --- achieves state-of-the-art results on two few-shot learning benchmarks. The key to the success of their baseline is using cosine classification function and applying standard data augmentation techniques during few-shot training. Here we confirm their observation about the surprising efficiency of this baseline in a more realistic setting and demonstrate that learning a classifier on top of the compositional feature representation results in a significant improvement in performance.%the authors of

{\bf Compositional representations} have been extensively studied in the cognitive science literature~\cite{biederman1987recognition,hoffman1984parts,fodor1975language}, with Biderman's Recognition-By-Components theory being especially influential in computer vision. One attractive property of compositional representations is that they allow learning novel concepts from a few or even a single example by composing known primitives. Lake \etal~\cite{lake2017building} argue that compositionality is one of the key building blocks of human intelligence that is missing in the state-of-the-art artificial intelligence systems. Although early computer vision models have been inherently compositional~\cite{marr1982vision,marr1978representation,winston1970learning}, building upon feature hierarchies~\cite{fidler2007towards,zhu2010part} and part-based models~\cite{ott2011shared,felzenszwalb2010object}, modern deep learning systems~\cite{lecun1995convolutional,krizhevsky2012imagenet,he2016identity} do not explicitly model concepts as combinations of parts.

Analysis of internal representations learned by deep networks~\cite{zeiler2014visualizing,simonyan2013deep,mahendran2015understanding,zhou2018interpreting,krishnan2016tinkering} has shown that some of the neurons in the hidden layers do encode object and scene parts. However, all the work observes that the discovered compositional structure is limited and qualitative analysis of network activations is highly subjective. Very recently, an approach to quantitative evaluation of compositionality of learned representations has been proposed by Andreas~\cite{x2018comp}. This work posits that a feature encoding of an image is compositional if it can be represented as a sum of the encodings of attributes describing the image, and designs an algorithm to quantify this property. We demonstrate that na\"{\i}vely turning this measure into a training objective results in inferior performance and we propose a remedy.%all these works an approach to  their

Among prior work that explicitly addresses compositionality in deep learning models, Misra \etal~\cite{misra2017red} propose to train a network that predicts classifiers for novel concepts by composing existing classifiers for the parts. 
%For instance, their model can obtain a classifier for the category {\tt large elephant} by combining independently learned classifiers for categories {\tt elephant} and {\tt large} without seeing a single image with a label {\tt large elephant}. 
By contrast, we propose to train a single model that {\em internally} decomposes concepts into parts and show results in a few-shot setting. Stone \etal~\cite{stone2017teaching} address the notion of spatial compositionality, constraining network representations of objects in an image to be independent from each other and from the background. They then demonstrate that networks trained with this constraint generalize better to the test distribution. While we also enforce decomposition of a network representation into parts with the goal of increasing its generalization abilities, our approach does not require spatial, or even image-level supervision. We can thus handle abstract attributes and be readily applied to large-scale datasets.
%Our work is similar in that we also enforce decomposition of a network representation into parts with the goal of increasing its generalization abilities. Our approach, however, does not require spatial, or even image-level supervision, and thus can handle abstract attributes, as well as being readily applied to large-scale datasets.%the authors propose to 

{\bf Learning with attributes} has been studied in a variety of applications. Most notably, zero-shot learning methods use category-level attributes to recognize novel classes without seeing any training examples~\cite{akata2013label,almazan2014word,deng2014large,lampert2009learning}. To this end, they learn models that take attributes as input and predict image classifiers, allowing them to recognize never-before-seen classes as long as they can be described by the known attribute vocabulary. 
%Very recently it has ben shown~\cite{x2018lll} that such attribute-based classifiers also require less data for training.
By contrast, our method uses attributes to learn compositional image representations that require fewer training examples to recognize novel concepts. Crucially, unlike these methods, our approach {\em does not require attribute annotations for novel classes}.%propose to described above

Another context in which attributes have been used is that of active~\cite{parkash2012attributes} and semi-supervised learning~\cite{shrivastava2012constrained}. In~\cite{parkash2012attributes} attribute classifiers are used to mine hard negative images for a category based on user feedback. %In~\cite{dasgupta2018learning} a method that explicitly constructs classifiers by combing discriminative attributes provided by the user into DNF formulas is proposed. the authors use the 
Our method is offline and does not require user interactions. In~\cite{shrivastava2012constrained} attributes are used to explicitly provide constraints when learning from a small number of labeled and a large number of unlabeled images. Our approach uses attributes to regularize a learned deep image representation, resulting in these constraints being implicitly encoded by the network.

\section{Our Approach}
\label{sec:method}
\subsection{Problem Formulation}
We consider the task of few-shot image classification. We have a set of base categories $\mathcal{C}_{base}$ and a corresponding dataset $S_{base}=\left\{ \left(x_{i},y_{i}\right), x_i \in \mathcal{X}, y_{i}\in\mathcal{C}_{base}\right\}$ which contains a large number of examples per class. We also have a set of unseen novel categories $\mathcal{C}_{novel}$ and a corresponding dataset $S_{novel}=\left\{ \left(x_{i},y_{i}\right), x_i \in \mathcal{X}, y_{i}\in\mathcal{C}_{novel}\right\}$ which consists of only $n$ examples per class, where $n$ could be as few as one. We learn a representation model $f_\theta$ parametrized by $\theta$ on $S_{base}$ that can be used for the downstream classification task on $S_{novel}$.%category Formally, w

While there might exist many possible representations that can be learned and achieve similar generalization performance on the base categories, we argue that the one that is decomposable into shared parts will be able to generalize better to novel categories from fewer examples. Consider again the example in Figure~\ref{fig:intro}. Intuitively, a model that has internally learned to recognize the attributes {\tt beak:curvy}, {\tt wing\_color:grey}, and {\tt breast\_color:white} is able to obtain a classifier of the never-before-seen bird species simply by composition. But how can this intuitive notion of compositionality be formulated in the space of deep representation models?

\begin{figure*}[t]
\begin{center}
\threefigures{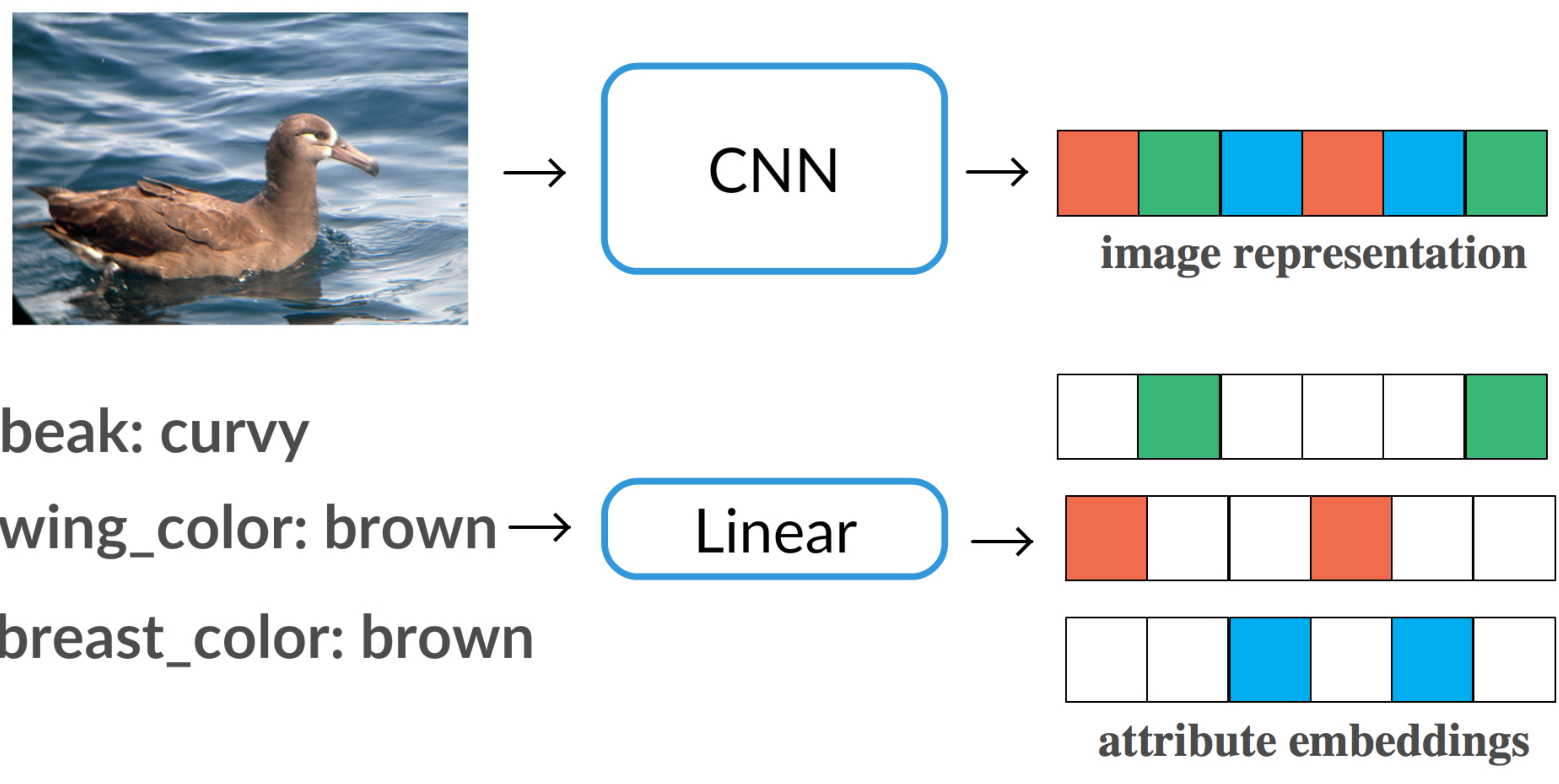}{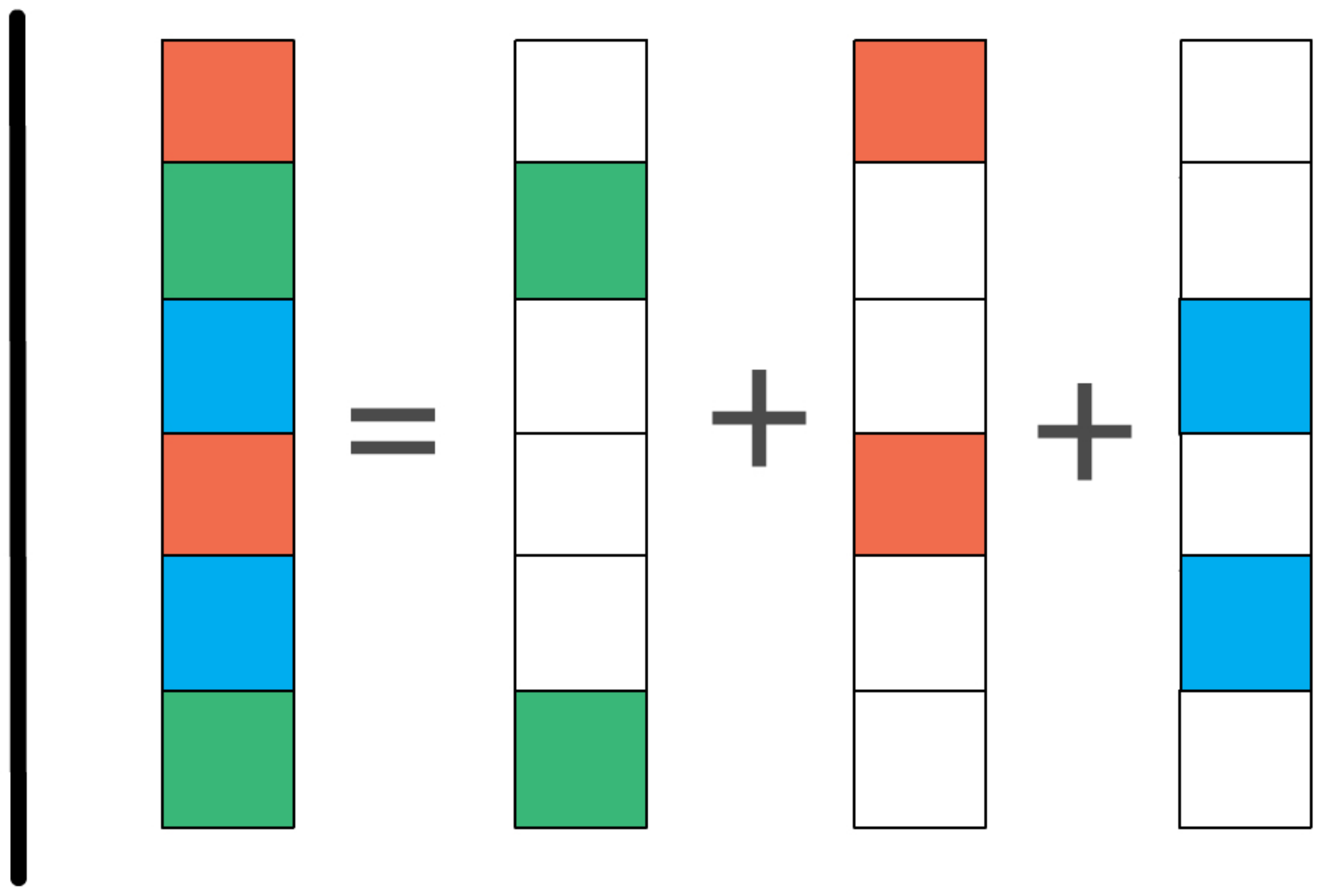}{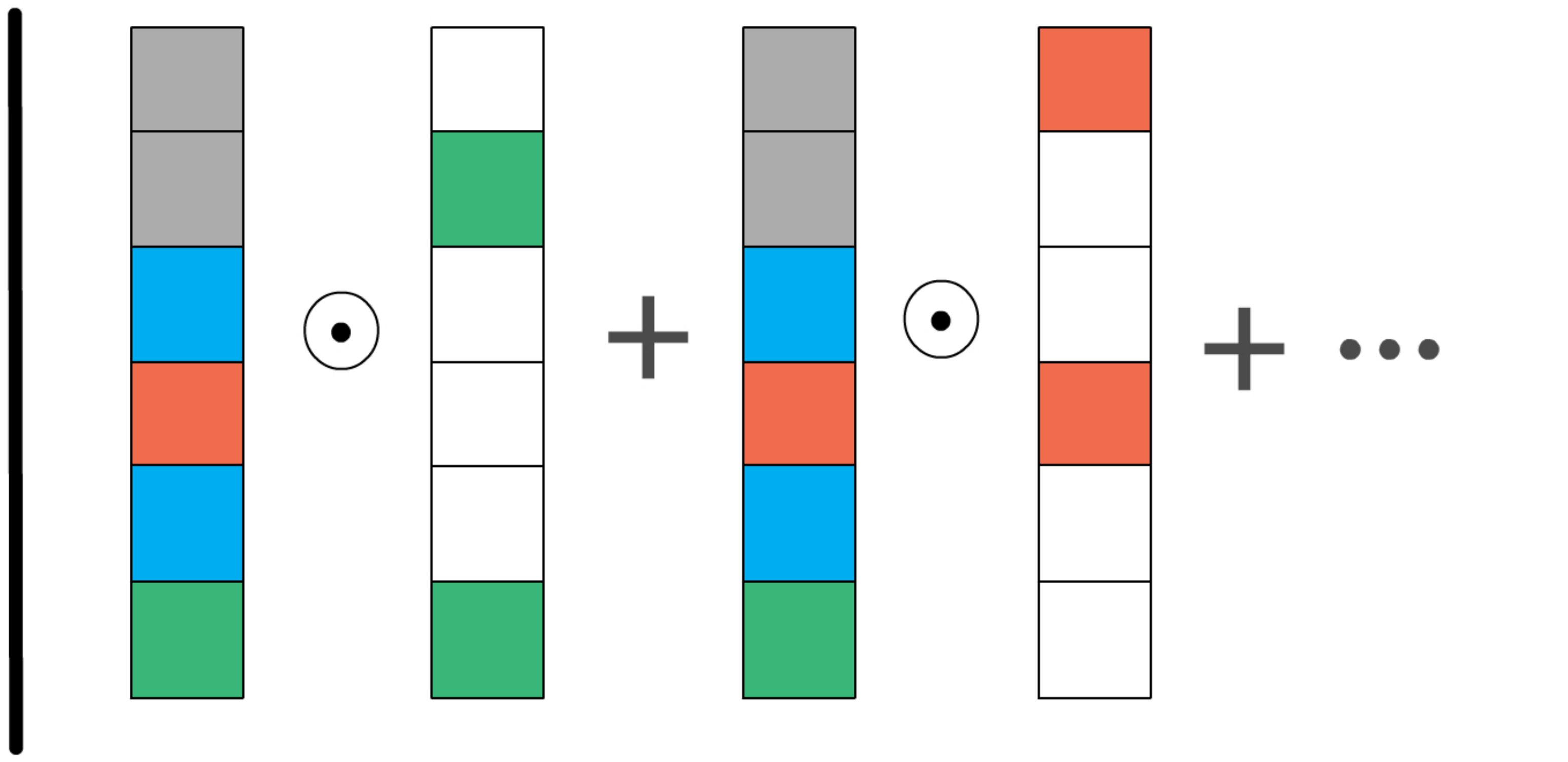}{0.35\linewidth} {0.24\linewidth} {0.33\linewidth} \vspace{0.1cm}
\end{center}
\vspace{-0.3cm}
\caption{Overview of our proposed compositional regularization. The goal is to learn an image representation that is decomposable into parts by leveraging attribute annotations. First, an image is encoded with a CNN and its attributes with a linear layer (a). We then propose two forms of regularizations: a hard one, shown in (b) and a soft one, shown in (c). The former is forcing the image representation to be fully described by the attributes. The latter is a relaxed version that allows a part of the representation to encode other information about the images (shown in grey).}
\label{fig:method}
\vspace{-0.5cm}
\end{figure*}

Following the formalism proposed in~\cite{x2018comp}, on the base dataset $S_{base}$, we augment the category labels $y_i \in \mathcal{C}_{base}$ of the examples $x_i$, with information about their structure in the form of {\em derivations} $D(x_i)$, defined over a set of {\em primitives} $\mathcal{D}_0$. That is, $D(x_i)$ is a subset of $\mathcal{D}_0$. In practice, these primitives can be seen as parts, or, more broadly, attributes capturing the compositional structure of the examples. Derivations are then simply sets of attribute labels. For instance, for the CUB-200-2011 dataset the set of primitives consists of items such as {\tt beak:curvy}, {\tt beak:needle}, \etc, and a derivation for the image in Figure~\ref{fig:intro}(a) is then \{{\tt beak:curvy}, {\tt wing\_color:brown}, ...\}.

We now leverage derivations to learn a compositional representation on the base categories. Note that for the novel categories, we only have access to the category labels {\em without} any derivations. 

\subsection{Compositionality Regularization}
In~\cite{x2018comp} a representation $f_{\theta}$ is defined as compositional over $\mathcal{D}_0$ if each $f_{\theta}(x)$ is determined by $D(x)$. That is, the image representation can be reconstructed from the representations of the corresponding attributes. This definition is formalized in the following way:
\begin{equation}
f_{\theta}(x_i) = \sum_{d \in D(x_i)} \hat{f}_{\eta}(d),
\label{eq:comp}
\end{equation}
where $\hat{f}_{\eta}$ is the attribute representation parameterized by $\eta$, and $d$ is an element of the derivation of $x_i$. In practice, $\hat{f}_{\eta}$ is implemented as a linear embedding layer (see Figure~\ref{fig:method}(a)), so $\eta$ is a matrix of size $k \times m$, where $k=| \mathcal{D}_0 |$, and $m$ is the dimensionality of the image embedding space. Given a fixed, pre-trained image embedding $f_{\theta}$, Eq.~\eqref{eq:comp} can be optimized over $\eta$ to discover the best possible decomposition. In~\cite{x2018comp} this decomposition is then used to evaluate a reconstruction error on a held-out set of images and quantify the compositionality of $f_{\theta}$.

By contrast, in this work we want to use attribute annotations to improve the compositional properties of image representations. Na\"{\i}vely, one could imagine a method that directly enforces the equality in Eq.~\eqref{eq:comp} while learning the image representation. Indeed, it is differentiable not only with respect to $\eta$ but also with respect to $\theta$. We can thus turn it into an objective function $\sigma(f_\theta(x_i), \sum_{d \in D(x_i)} \hat{f}_{\eta}(d))$, where $\sigma$ is a distance function, such as cosine similarity, and jointly optimize both ${f_\theta}$ and $\hat{f}_\eta$. 
\begin{figure*}[t]
\begin{center}
\includegraphics[trim={1cm 2.8cm .6cm 0},clip,width=\linewidth]{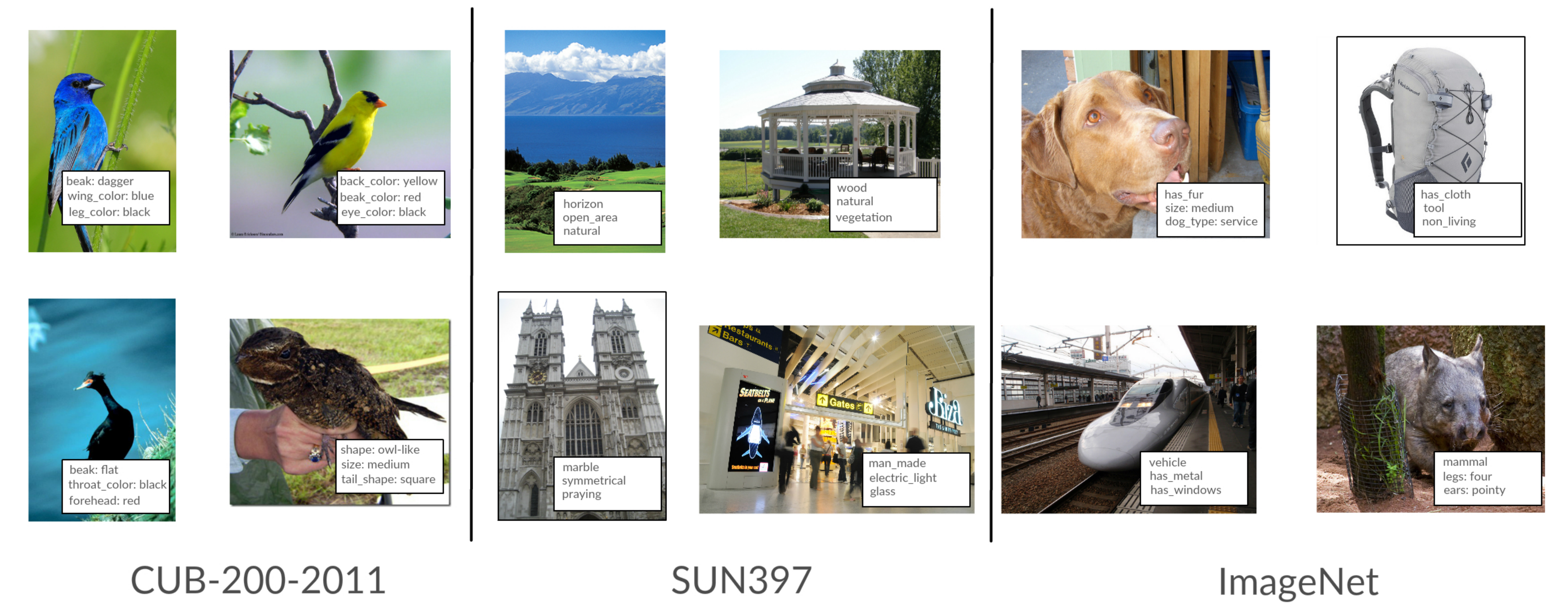}\\
{\small \hspace{-1cm}\text{CUB-200-2011} \hspace{3.5cm} \text{ SUN397}\hspace{5cm} \text{ImageNet}}
 \vspace{0.25cm}
\end{center}
\vspace{-.6cm}
\caption{Examples of categories from three datasets used in the paper together with samples of attribute annotations.}
\label{fig:data}
\vspace{-0.5cm}
\end{figure*}

\textbf{Hard constraints:} Based on this observation, we propose a hard compositionality constraint:
\begin{equation}
L_{cmp\_h}(\theta,\eta) = \sum_{i} \sigma\Big(f_\theta(x_i), \sum_{d \in D(x_i)} \hat{f}_{\eta}(d)\Big).
\label{eq_ltre}
\end{equation}
It can be applied as a regularization term together with a classification loss $L_{cls}$, such as softmax. Intuitively, Eq.~\eqref{eq_ltre} imposes a constraint on the gradient-based optimization of parameters $\theta$, forcing it to choose out of all the representations that solve the classification problem equally well the one that is \textit{fully decomposable} over a pre-defined vocabulary of primitives $\mathcal{D}_0$. A visualization of the hard constraint is presented in Figure~\ref{fig:method}(b). Overall, we use the following loss for training:
\begin{equation}
L(\theta,\eta) = L_{cls}(\theta) + \lambda L_{cmp\_h}(\theta,\eta),
\label{eq:loss_full}
\end{equation}
where $\lambda$ is a hyper-parameter that balances the importance of the two objectives.

One crucial assumption made in Eq.~\eqref{eq:comp} is that the derivations $D$ are exhaustive. In other words, for this equation to hold, $\mathcal{D}_0$ has to capture all the aspects of the images that are important for the downstream classification task. However, even in such a narrow domain as that of CUB, exhaustive attribute annotations are extremely expensive to obtain. In fact, it is practically impossible for larger-scale datasets, such as SUN and ImageNet. Ideally, we want only a part of the image embedding $f_{\theta}$ to model the primitives in $\mathcal{D}_0$, allowing the other part to model the remaining factors of variation in the data. More formally, we want to enforce a softer constraint compared to the one in Eq.~\eqref{eq:comp}: 
\begin{equation}
f_\theta(x_i)= \sum_{d \in D(x_i)} \hat{f}_\eta(d) + w(x_i), 
\label{eq:obj_soft}
\end{equation}
where $w(x_i)$ accounts for a part of the image representation which is not described by the attributes. 

To this end, instead of enforcing the full decomposition of the image embedding over the attributes, we propose to maximize the sum of the individual similarities between the embedding of each attribute and the image embedding using the dot product: $\sum_{d \in D(x_i)} f_{\theta}(x_i) \cdot \hat{f}_{\eta}(d)$. Optimizing this objective jointly with $L_{cls}$ ensures that $f_{\theta}$ captures the compositional information encoded by the attributes, while allowing it to model the remaining factors of variation that are useful for the classification task. Note that to avoid trivial solutions, the similarity with the embeddings of the attributes that are not in $D(x_i)$ has to be minimized at the same time.%each attribute's embedding

\textbf{Soft constraints:} Our proposed soft compositionality constraint is defined below (see Figure~\ref{fig:method}(c) for a visualization):
\begin{equation}
\begin{split}
& L_{cmp\_s}(\theta,\eta) =  \\
& \sum_{d \not \in D(x_i)} f(x_i) \cdot \hat{f}_\eta(d) - \sum_{d \in D(x_i)} f(x_i) \cdot \hat{f}_\eta(d).
\label{eq:stre}
\end{split}
\end{equation}
It is easy to see that this formulation is equivalent to multi-label classification when weights of attribute classifiers are treated as embeddings $\hat{f}_\eta(d)$. In contrast to the hard variant in Eq.~\eqref{eq_ltre}, it allows a part of the image encoding $f_{\theta}$, shown in grey in Figure~\ref{fig:method}(c)), to represent the information not captured by the attribute annotations. %In Section~\ref{sec:anal} we empirically verify that this relaxation is critical for obtaining top performance.

Finally, we observe that some attributes can be highly correlated in the training set. For instance, most of the {\tt natural} scenes on SUN also have {\tt vegetation} in them. Directly optimizing Eq.~\eqref{eq:stre} for such attributes will fail to disentangle the corresponding factors of variation, limiting the generalization abilities of the learned image representation. To address this issue, we propose to enforce orthogonality of the attribute embeddings $\hat{f}_{\eta}$. In particular, our final objective takes the form:
\begin{equation}
L(\theta,\eta) = L_{cls}(\theta) + \lambda L_{cmp\_s}(\theta,\eta) + \beta \mid \eta \eta^T - I \mid,
\label{eq:orth}
\end{equation}
where $I$ is the identity matrix, and $\beta$ is a hyper-parameter that controls the importance of the orthogonality constraint. Notice that similar constraints have been recently proposed in other domains~\cite{bansal2018can,brock2018large}.

\subsection{Complexity of Obtaining Attribute Supervision}
\label{sec:imnet_attrs}
Up till now the derivations $D(x_i)$ were defined on the instance level. Such fine-grained supervision is very expensive to obtain. To mitigate this issue, we claim that instances in any given category share the same compositional structure. Indeed, all seagulls have curvy beaks and short necks, so we can significantly reduce the annotation effort by redefining derivations as $D(x_i) = D(y_{i})$. One objection might be that the beak is not visible in all the images of seagulls. While this is true, we argue that such labeling noise can be ignored in practice, which is verified empirically in Section~\ref{sec:exp}.

We use three datasets for experimental evaluation: CUB-200-2011~\cite{wah2011caltech}, SUN397~\cite{xiao2010sun}, and ImageNet~\cite{deng2009imagenet}. Samples of images from different categories of the three datasets, together with their attribute annotations, are shown in Figure~\ref{fig:data}. Our method handles concrete visual attributes, like {\tt material} and {\tt color}, and abstract attributes, such as {\tt openness} and {\tt symmetry}. For the first two datasets, attribute annotations are publicly available, but for ImageNet we collect them ourselves. Below we describe key steps in collecting these annotations. 

We heavily rely on the WordNet~\cite{miller1995wordnet} hierarchy both to define the vocabulary of attributes and to collect them. First, we define attributes on each level of the hierarchy: every object has {\tt size} and {\tt material}, most of the mammals have {\tt legs} and {\tt eyes}, \etc This allows us to obtain a vocabulary that is both broad, intersecting boundaries of categories, and specific enough, capturing discriminative properties. Second, we also rely on the hierarchical properties of the attributes to simplify annotation process (see Appendix~\ref{sec:attr}). In particular, the annotator is first asked about generic properties of the category, like whether it is {\tt living}, and then all the attributes specific to {\tt non-living} entities are set to a negative value automatically. This pruning is applied on every level of the hierarchy, allowing a single annotator to collect attribute labels for 386 categories in the base split of~\cite{hariharan2017low} in just 3 days.

\section{Experiments}\label{sec:experiments}
\label{sec:exp}

\subsection{Datasets and Evaluation}
\label{sec:data}
We use three datasets: CUB-200-2011, SUN397, and ImageNet. 
%For the first two datasets we employ attribute annotations collected externally. For ImageNet no prior attribute annotations exist, so we collect them ourselves. for experimental analysis
Below we describe each of them together with their evaluation protocols in more detail.%

\textbf{CUB-200-2011} is a dataset for fine-grained classification~\cite{wah2011caltech}. It contains 11,788 images of birds, labeled with 200 categories corresponding to bird species. The dataset is evenly split into training and test subsets. In addition, the authors have collected annotations for 307 attributes, corresponding to the appearance of the birds' parts, such as shape of the beak and color of the forehead. These attribute annotations have been collected on the image level via crowd sourcing. We {\em aggregate them on the category level} by labeling a category as having a certain attribute if at least half of the images in the category are labeled with it. We further filter out rare attributes by only keeping the ones that are labeled for at least five categories, resulting in 130 attributes used in training. For few-shot evaluation, we randomly split the 200 categories into 100 base and 100 novel categories.

\textbf{SUN397} is a subset of the SUN dataset for scene recognition, which contains the 397 most well sampled categories, totaling to 108,754 images~\cite{xiao2010sun}. Patterson \etal~\cite{patterson2014sun} have collected discriminative attributes for these scene categories,
%In particular, each annotator has been shown four images representative of four random categories and then asked to name attributes that distinguish them from the other two. This 
resulting in a vocabulary of 106 attributes that are both discriminative and shared across scene classes. 
%Examples include both abstract attributes, such as `man-made', and purely visual ones, such as `grass'. 
Similar to CUB, we aggregate these image-level annotations for categories by labeling a category as having an attribute if half of the images in the category have this attribute, and filter out the infrequent attributes, resulting in 89 attributes used in training. For few-shot evaluation, we randomly split the scene categories into 197 base and 200 novel categories.

\textbf{ImageNet} is an object-centric dataset~\cite{deng2009imagenet} that contains 1,200,000 images labeled with 1,000 categories. The categories are sampled from the WordNet~\cite{miller1995wordnet} hierarchy and constitute a diverse vocabulary of concepts ranging from animals to music instruments. Defining a vocabulary of attributes for such a dataset is non-trivial and has not been done previously. We described our approach for collecting the attributes in more detail in Section~\ref{sec:imnet_attrs}. For few-shot evaluation, we use the split proposed in~\cite{hariharan2017low,wang2018low}.

\subsection{Implementation Details}
Following~\cite{hariharan2017low,wang2018low}, we use a ResNet-10~\cite{he2016identity} architecture as a backbone for all the models, but also report results using deeper variants in Appendix~\ref{sec:depth}. We add a linear layer without nonlinearity at the end of all the networks to aid in learning a cosine classifier. The networks are first pre-trained on the base categories using mini-batch SGD, as in~\cite{hariharan2017low,wang2018low,x2018fewshot}. The learning rate is set to 0.1, momentum to 0.9, and weight decay to 0.0001. The batch size and learning rate schedule depend on the dataset size and are reported in Appendix~\ref{sec:impl}. 
%In particular, for ImageNet and SUN397, we use the setting proposed in~\cite{hariharan2017low,wang2018low} with a batch size of 256 and 90 training epochs. The learning rate is decreased by a factor of 10 every 30 epochs. For CUB-200-2011, which is a much smaller dataset, we use a batch size of 16 and train for 170 epochs. The learning rate is first decreased by a factor of 10 after 130 epochs, and then again after 20 more epochs. This schedule is selected on the validation set. We use both linear and cosine classifiers proposed in~\cite{gidaris2018dynamic,x2018fewshot} in the experiments. 
All the models are trained with a sofmax cross-entropy loss as $L_{cls}$ in Eq.~\eqref{eq:loss_full}. Our soft compositionality constraint in Eq.~\eqref{eq:stre} is implemented with a multi-label, one-versus-all loss.

We observe that the proposed compositionality regularization slows down convergence when training from scratch. To mitigate this issue, we first pre-train a network with the standard classification loss and then fine-tune it with regularization for the same number of epochs using the same optimization parameters. For a fair comparison, baseline models are fine-tuned in the same way. We set the hyper-parameters $\lambda$ and $\beta$ in Eq.~\eqref{eq:orth} for each dataset individually, using the validation set.
%For ImageNet and SUN397 $\lambda$ is set to 8, and for CUB-200-2011 to 10. The attribute annotations are sparse, with around 10\% of them being labeled as positive for any given image on average. 
%Due to this highly imbalanced distribution of training labels, all the attribute classifiers learn to predict the negative labels. To address it, 
%We randomly sample a subset of the negative attributes for every example in every batch to balance the number of positive and negative examples.

In few-shot training, we use the baseline proposed in~\cite{x2018fewshot} as our base model. In particular, we learn either a linear or cosine classifier on top of the frozen CNN representation. Differently from~\cite{x2018fewshot}, we learn the classifier jointly on novel and base categories. We use mini-batch SGD with a batch size of 1,000 and a learning rate of 0.1, but find that training is robust to these hyper-parameters. What is important is the number of training iterations. This number depends on the dataset and the classifier (see Appendix~\ref{sec:impl}). 
%On ImageNet we train for 100 iterations for both Cosine and linear classifiers. On SUN397 we train for 200 iterations for linear and 100 for Cosine classifier. On CUB we train for 100 iterations for linear and 40 for Cosine classifier. To select these values, we split the base categories in half, using one half as validation, and train until the top-5 performance on the validation categories stops increasing. We use the same setting to select the optimal hyper-parameters for other methods. 
Overall, we follow the evaluation protocol proposed in~\cite{wang2018low}.
\begin{figure}[t!]
\centering
\includegraphics[height=.6\linewidth]{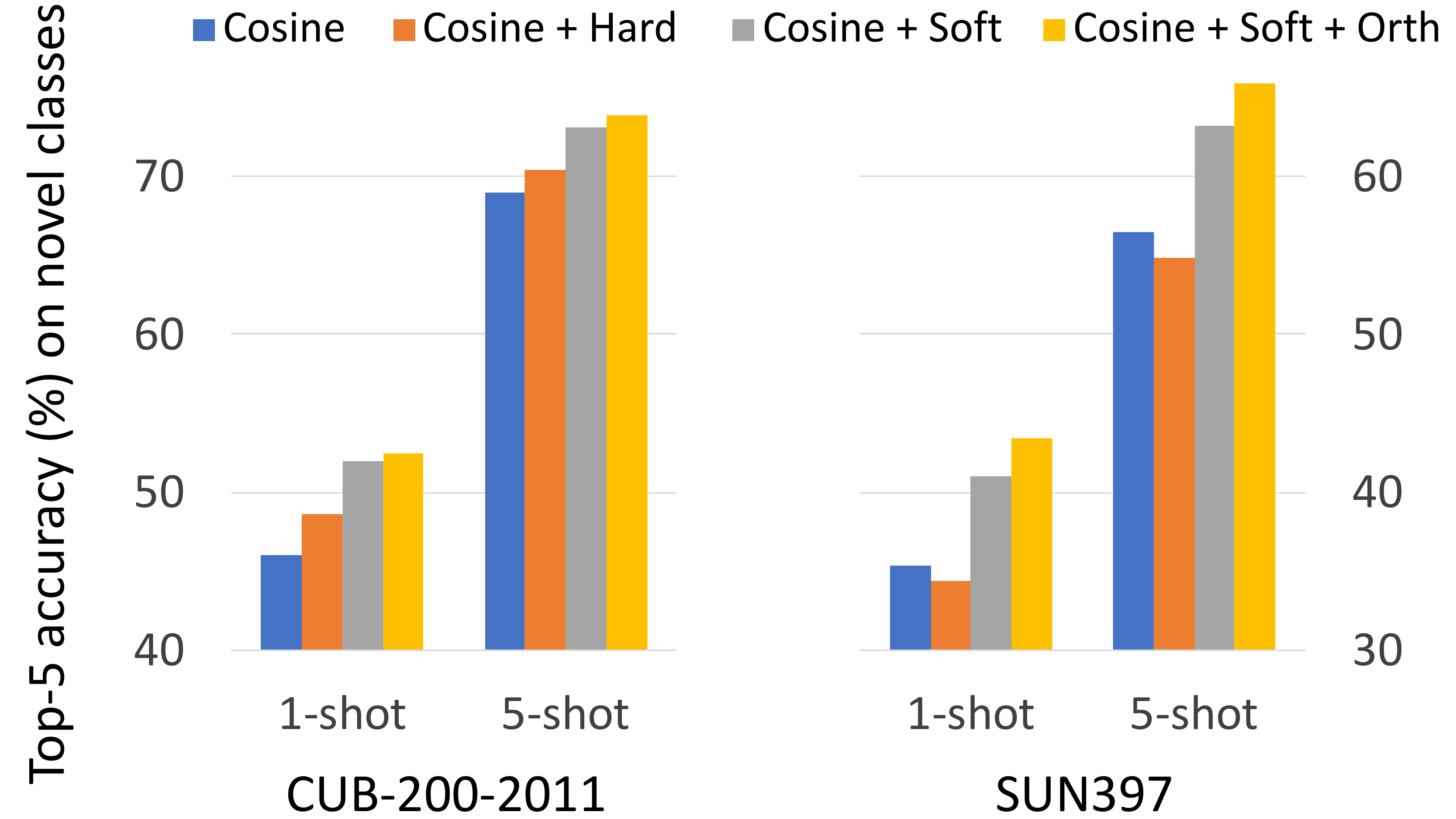}
\caption{Comparison of the variants of our compositionallity regularizations to a baseline on the novel categories of the CUB and SUN datasets. The y-axis indicates top-5 accuracy. Our soft regularization with orthogonality constraint achieves the best performance on both datasets.}
\label{fig:cosine_bar}
\vspace{-5mm}
\end{figure}

\subsection{Analysis of Compositional Representations}
\label{sec:anal}
Here we analyze whether our compositionallity constraints lead to learning representations that are able to recognize novel categories from a few examples. We use CUB and SUN datasets due to availability of high quality annotations. Following~\cite{gidaris2018dynamic,x2018fewshot}, we use a cosine classifier in the most of the experiments due to its superior performance. A qualitative analysis of the representations using Network Dissection~\cite{zhou2018interpreting} is provided in Appendix~\ref{sec:netdissect}.%In this section, w proposed in Section~\ref{sec:method} also 

\textbf{Comparison between hard and soft compositional constraints:}
We begin our analysis by comparing the two proposed variants of compositionallity regularizations: the hard one in Eq.~\eqref{eq_ltre} and the soft one in Eq.~\eqref{eq:stre}. Figure~\ref{fig:cosine_bar} shows the top-5 performance on the novel categories of CUB and SUN. We perform the evaluation in 1- and 5-shot scenarios. First we notice that the variant of the regularization based on the hard sum constraint (shown in orange) slightly increases the performance over the baseline on CUB, but leads to a decrease on SUN. This is not surprising, since this constraint assumes exhaustive attribute annotations, as mentioned in Section~\ref{sec:method}. On CUB the annotations of bird parts do capture most of the important factors of variation, whereas the attributes on SUN are less exhaustive. By contrast, our proposed soft constraint (shown in grey) allows the representations to capture important information that is not described in the attribute annotation. Enforcing orthogonality of the attribute embeddings, as proposed in Eq.~\eqref{eq:orth}, further improves the performance. The improvement is larger on SUN, since attributes are more correlated in this dataset. Overall, our full model (shown in yellow) improves the performance over the baseline by 6.4\% on CUB and by 8\% on SUN in the most challenging 1-shot scenario. Comparable improvements are observed in the 5-shot regime as well. This confirms our hypothesis that enforcing the learned representation to be decomposable over category-level attributes allows it to generalize to novel categories with fewer examples. We thus use the soft variant of our approach with orthogonality constraint in the remainder of the paper. 

%Finally, the improvement of our compositional model over the baseline decreases slightly on CUB, as it sees more examples from the novel categories. This is because, as the training regime gets closer to the standard data-rich scenario, additional regularization methods become redundant. Nevertheless, our method still improves over the baseline by 5.7\%. On SUN, the improvement actually increases to 9.5\%.
\begin{table}[bt]
 \centering
  {
    \footnotesize
\resizebox{\linewidth}{!}{
    \begin{tabular}{l|c@{\hspace{1em}}c@{\hspace{1em}}c@{\hspace{1em}}|c@{\hspace{1em}}c@{\hspace{1em}}c@{\hspace{1em}}c}
      & \multicolumn{3}{c|}{Novel} &
          \multicolumn{3}{c}{All}\\
        & 1-shot          & 2-shot           & 5-shot      
        & 1-shot           & 2-shot            & 5-shot         \\\hline
    Cos
        & 46.1       & 57.0       & 68.9    
        & 58.2       & 63.3       & 69.8     \\
    Cos w/ comp
        & 52.5       & 63.6       & 73.8     
        & 62.6       & 68.4       & 74.0       \\
    Linear w/ comp
        & 47.0       & 60.0       & 74.0     
        & 57.6       & 65.0       & 72.7        \\\hline
     Cos + data aug
        & 47.7       & 58.0       & 69.4    
        & 58.7       & 64.0       & 70.1     \\
    Cos w/ comp + data aug
        & \textbf{53.6}       & \bf{64.8}       & 74.6   
        & \textbf{63.1}     & \textbf{69.2}      & \bf{74.5}      \\
    Linear w/ comp + data aug
        & 52.1       & 63.1       & \bf{75.2}    
        & 60.9       & 67.3       & 73.9    \\
\end{tabular}
}
}
\caption{Analysis of our approach: top-5 accuracy on the novel and all (\ie, $novel+base$) categories of the CUB dataset. `Cos': the baseline with a cosine classifier, `Cos w/ comp': our compositional representation with a cosine classifier, `Linear w/ comp': our compositional representation with a linear classifier. The variants trained with data augmentation are marked with `+ data aug'.}
\vspace{-2mm}
\label{tab:anal_cub}
\end{table}
\begin{table}[bt]
 \centering
  {
\resizebox{\linewidth}{!}{
    \begin{tabular}{l|c@{\hspace{1em}}c@{\hspace{1em}}c@{\hspace{1em}}|c@{\hspace{1em}}c@{\hspace{1em}}c@{\hspace{1em}}c}
      & \multicolumn{3}{c|}{Novel} &
          \multicolumn{3}{c}{All}\\
        & 1-shot          & 2-shot           & 5-shot      
        & 1-shot           & 2-shot            & 5-shot         \\\hline
    Cos
        & 35.4       & 45.6       & 56.4    
        & 52.1       & 56.7       & 61.9     \\
    Cos w/ comp
        & 43.4       & 54.5       & 65.9      
        & 54.9       & 60.4       & 66.3       \\
    Linear w/ comp
        & 41.2       & 51.8       & 63.3     
        & 50.1       & 57.6       & 66.4        \\\hline
     Cos + data aug
        & 39.9       & 49.7       & 59.7    
        & 54.2       & 58.5       & 63.5     \\
    Cos w/ comp + data aug
        & \bf{45.9}       & \bf{56.7}       & \bf{67.1}    
        & \bf{56.3}       & \bf{61.5}      & \bf{67.3}      \\
    Linear w/ comp + data aug
        & 41.1       & 51.6       & 62.3    
        & 51.7       & 57.1       &  63.3   \\
\end{tabular}
}
}
\caption{Analysis of our approach: top-5 accuracy on the novel and all (\ie, $novel+base$) categories of the SUN dataset. `Cos': the baseline with a cosine classifier, `Cos w/ comp': our compositional representation with a cosine classifier, `Linear w/ comp': our compositional representation with a linear classifier. The variants trained with data augmentation are marked with `+ data aug'.}
\vspace{-2mm}
\label{tab:anal_sun}
\end{table}

\textbf{Ablation studies:}
We further analyze the compositional representation learned with the soft constraint through extensive ablations and report the results in Table~\ref{tab:anal_cub}. 

\emph{Evaluation in the challenging joint label space of base and novel classes:}
We notice that the observation about the positive effect of the compositionallity constraints on the generalization performance of the learned representation made above for the novel categories holds for the $novel+base$ setting (right part `All' of Table~\ref{tab:anal_cub}, rows 1 and 2). In particular, our approach improves over the baseline by 4.4\% in the 1-shot and by 4.2\% in the 5-shot setting. 

\emph{Cosine vs. linear classifiers:}
The linear classifier, denoted as `Linear w/ comp', performs significantly worse than the cosine variant, especially in the $novel+base$ setting. A similar behavior was observed in~\cite{gidaris2018dynamic,x2018fewshot} and attributed to that the cosine classifier explicitly reduces intra-class variation among features during training, by unit-normalizing the vectors before dot product operation.
\begin{figure}[t!]
\centering
\includegraphics[trim=0 0 8 0, clip, height=.6\linewidth]{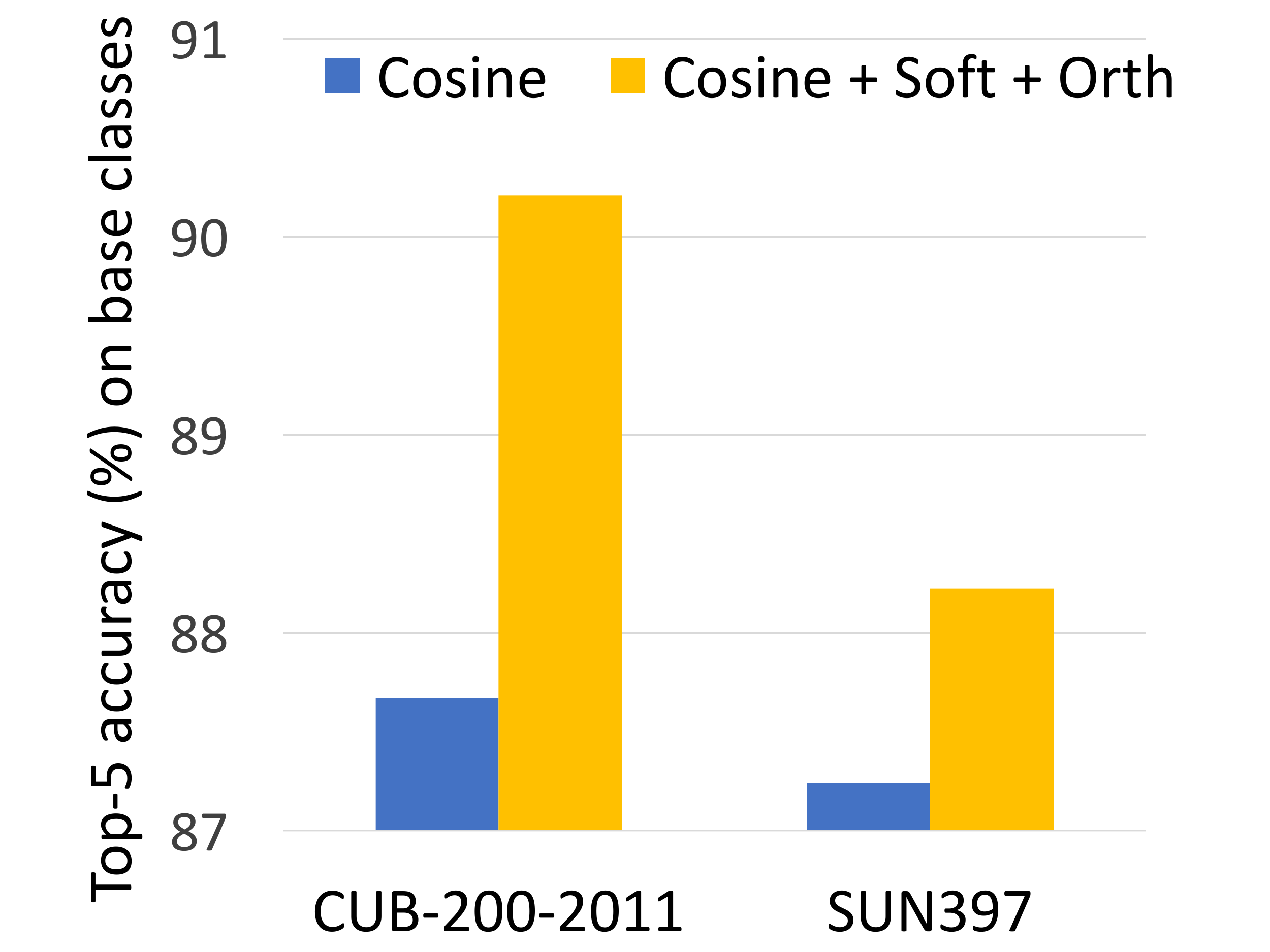}
\caption{Comparison of our compositionallity regularization to a baseline on the base categories of the CUB and SUN datasets. The y-axis indicates top-5 accuracy on the corresponding validation set. The improvements are smaller than those on the novel classes.}
\label{fig:cosine_base_bar}
\vspace{-4mm}
\end{figure}

\emph{Effect of data augmentation:}
Another observation made in~\cite{x2018fewshot} is that, for a fair comparison, standard data augmentation techniques, \eg, random cropping and flipping, need to be applied when performing few-shot learning. We report the results with data augmentation in the lower part of Table~\ref{tab:anal_cub}. The most important result here is that data augmentation is indeed effective when learning a classifier in a few-shot regime, improving the performance of all the variants. By contrast, in Appendix~\ref{sec:exp_add} we demonstrate that traditional few-shot learning methods are not able to significantly benefit from data augmentation.
\begin{table*}[t]
\begin{center}
\resizebox{.86\linewidth}{!}{
    \begin{tabular}{l|c@{\hspace{1em}}c@{\hspace{1em}}c@{\hspace{1em}}c@{\hspace{1em}}|c@{\hspace{1em}}c@{\hspace{1em}}c@{\hspace{1em}}c}
      & \multicolumn{4}{c|}{Novel} &
          \multicolumn{4}{c}{All}\\
        & 1-shot          & 2-shot           & 5-shot  & 10-shot      
        & 1-shot           & 2-shot            & 5-shot  & 10-shot         \\\hline
    Prototypical networks~\cite{snell2017prototypical}
        & 43.2       & 54.3       & 67.8 & 72.9    
        & 55.6       & 59.1       & 64.1 & 65.8     \\
    Matching Networks~\cite{vinyals2016matching}
         & 48.5       & 57.3       & 69.2 & 74.5    
        & 50.6      & 55.8       & 62.6 & 65.4     \\
    Relational networks~\cite{yang2018learning}
        & 39.5       & 54.1       & 67.1 & 72.7    
        & 51.9       & 57.4       & 63.1 & 65.3           \\\hline
    Cos w/ comp (Ours)
         & 52.5       & 63.6       & 73.8 & 78.5      
        & 62.6       & 68.4       & 74.0 &  76.4      \\
    Cos w/ comp + data aug (Ours)
        & \bf{53.6}       & \bf{64.8}       & \bf{74.6} & \bf{78.7}    
        & \bf{63.1}       & \bf{69.2}       & \bf{74.5} & \bf{76.9}      \\
\end{tabular}
}
\caption{Comparison to the state-of-the-art approaches: top-5 accuracy on the novel and all (\ie, $novel+base$) categories of the CUB dataset using a  ResNet-10 backbone. Our approach consistently achieves the best performance.}
\label{tab:soa_cub}
\vspace{-3mm}
\end{center}
\end{table*}
\begin{table*}[t]
\begin{center}
\resizebox{.86\linewidth}{!}{
    \begin{tabular}{l|c@{\hspace{1em}}c@{\hspace{1em}}c@{\hspace{1em}}c@{\hspace{1em}}|c@{\hspace{1em}}c@{\hspace{1em}}c@{\hspace{1em}}c}
      & \multicolumn{4}{c|}{Novel} &
          \multicolumn{4}{c}{All}\\
        & 1-shot          & 2-shot           & 5-shot  & 10-shot      
        & 1-shot           & 2-shot            & 5-shot  & 10-shot         \\\hline
    Prototypical networks~\cite{snell2017prototypical}
        & 37.1       & 49.2       & 63.1 & 70.0    
        & 51.3      & 59.0       & 66.4 & 69.3     \\
    Matching Networks~\cite{vinyals2016matching}
         & 41.0       & 48.9       & 60.4 & 67.6    
        & 50.3       & 54.0       & 60.2 & 64.4     \\
    Relational networks~\cite{yang2018learning}
        & 35.1       & 49.0       & 63.7 & 70.3    
        & 51.0       & 58.6       & 66.5 & 69.1          \\\hline
    Cos w/ comp (Ours)
        & 43.4       & 54.5       & 65.9 &  71.4    
        & 54.9       & 60.4       & 66.3  & 69.2    \\
    Cos w/ comp + data aug (Ours)
         & \bf{45.9}       & \bf{56.7}       & \bf{67.1}  & \bf{72.3}  
        & \bf{56.3}       & \bf{61.5}       & \bf{67.3} &  \bf{70.0}    \\
\end{tabular}
}
\caption{Comparison to the state-of-the-art approaches: top-5 accuracy on the novel and all (\ie, $novel+base$) categories of the SUN dataset using a ResNet-10 backbone. Our approach consistently achieves the best performance.}
\label{tab:soa_sun}
\end{center}
\vspace{-3mm}
\end{table*}
\begin{table*}[!htp]
\begin{center}
\resizebox{.9\linewidth}{!}{
    \begin{tabular}{l|c@{\hspace{1em}}c@{\hspace{1em}}c@{\hspace{1em}}c@{\hspace{1em}}|c@{\hspace{1em}}c@{\hspace{1em}}c@{\hspace{1em}}c}
      & \multicolumn{4}{c|}{Novel} &
          \multicolumn{4}{c}{All}\\
        & 1-shot          & 2-shot           & 5-shot  & 10-shot      
        & 1-shot           & 2-shot            & 5-shot  & 10-shot         \\\hline
    Prototype Matching Networks w/ G~\cite{wang2018low}
& 45.8 & 57.8 & 69.0 & \bf{74.3}    
& 57.6 & 64.7 & \bf{71.9} & \bf{75.2}     \\
    Prototype Matching Networks~\cite{wang2018low}
& 43.3 & 55.7 & 68.4 & 74.0 
& 55.8 & 63.1 & 71.1 & 75.0     \\
Prototypical networks w/ G~\cite{wang2018low}
& 45.0 & 55.9 & 67.3 & 73.0
& 56.9 & 63.2 & 70.6 & 74.5     \\
    Prototypical networks~\cite{snell2017prototypical}
& 39.3 & 54.4 & 66.3 & 71.2   
& 49.5 & 61.0 & 69.7 & 72.9     \\
    Matching Networks~\cite{vinyals2016matching}
& 43.6 & 54.0 & 66.0 & 72.5   
& 54.4 & 61.0 & 69.0 & 73.7     \\\hline
    Cos w/ comp (Ours)
        &  46.6     &    58.0    & 68.5 & 73.0    
        &   55.4     &   63.8     &  71.2  & 74.4    \\
    Cos w/ comp + data aug (Ours)
        &  \bf{49.3}    &   \bf{60.1}   & \bf{69.7} & 73.6    
        &   \bf{57.9}    &   \bf{65.3}    &  \bf{71.9}  & 74.7    \\
\end{tabular}
}
\caption{Comparison to the state-of-the-art approaches: top-5 accuracy on the novel and all (\ie, $novel+base$) categories of the ImageNet dataset using a ResNet-10 backbone. Even with {\em noisy or less discriminative} attributes which we collected, our approach achieves the best performance in 1-, 2-, and 5-shot scenarios. It can be potentially combined with the data generation approach~\cite{wang2018low} for further improvement.}
\vspace{-3mm}
\label{tab:soa_imagenet}
\end{center}
\end{table*}

\emph{Larger-scale evaluation:}
To validate our previous observations, we now report results on a much larger-scale SUN397 dataset~\cite{xiao2010sun}. Table~\ref{tab:anal_sun} summarizes the 200- and 397-way evaluation in $novel$ and $novel+base$ settings, respectively. Overall, similar conclusions can be drawn here, confirming the effectiveness of our approach across domains and dataset sizes.

\emph{Effect of the number of attributes:} We also evaluate the effect of the number of attributes used for training on the few-shot performance in Appendix~\ref{sec:attr_num}.

\textbf{Large sample performance:}
Figure~\ref{fig:cosine_base_bar} evaluates the accuracy of the cosine classifier baseline (shown in blue) and our compositional representations (shown in yellow) on the validation set of the base categories of CUB and SUN. The improvement due to compositional representations is significantly lower than that on the novel categories (\eg, in the 1-shot scenario: 2.5\% compared with 6.4\% on CUB, and only 1\% compared with 8\% on SUN). This observation confirms our claim that the proposed approach does not simply improve the overall performance of the model, but increases its generalization ability in the few-shot regime.

\subsection{Comparison to the State-of-the-Art}
We now compare our compositional representations with the cosine classifiers, denoted as `Cos w/ comp', to the state-of-the-art few-shot methods based on meta-learning~\cite{vinyals2016matching,snell2017prototypical,wang2018low,yang2018learning}. We evaluate on 3 datasets: CUB-200-2011, SUN397, and ImageNet. For CUB and SUN which have publicly available, well annotated attributes, Tables~\ref{tab:soa_cub} and~\ref{tab:soa_sun} show that our approach easily outperforms all the baselines across the board even {\em without} data augmentation (in Appendix~\ref{sec:exp_add}, we show that other methods demonstrate marginal improvements with data augmentation). In particular, our full method provides 5 to 7 point improvements on both datasets for the novel classes in the most challenging 1- and 2-shot scenarios, and achieves similar improvements in the joint label space.%almost no around 

Table~\ref{tab:soa_imagenet} summarizes the comparison on ImageNet for which we collected attribute annotations ourselves. Here we compare to the state-of-the-art methods reported in~\cite{wang2018low}, including the approaches that generate additional training examples. These results verify the effectiveness of our approach to annotating attributes on the category level. The collected annotations might be noisy or less discriminative, compared to the crowd sourced annotation in~\cite{wah2011caltech,patterson2014sun}. However, our compositional representation with a simple cosine classifier still achieves the best performance in 1-, 2-, and 5-shot scenarios, and is only outperformed in the 10-shot scenario by Prototype Matching Networks~\cite{wang2018low}.

\section{Conclusion}
In this work, we have proposed a simple attribute-based regularization approach that allows to learn compositional image representations. The resulting representations are on par with the existing  approaches when many training examples are available, but generalize much better in the small-sample size regime. We validated the use of our approach in the task of learning from few examples, obtaining the state-of-the-art results on three dataset. Compositionality is one of the key properties of human cognition that is missing in the modern deep learning methods, and we believe that our work is a precursor to a more in-depth study on this topic.

{\footnotesize {\bf Acknowledgments:} This work was supported in part by the Intelligence Advanced Research Projects Activity (IARPA) via
Department of Interior/Interior Business Center (DOI/IBC) contract number D17PC00345, ONR MURI N000141612007, and U.S. Army Research Laboratory (ARL) under the Collaborative Technology Alliance Program, Cooperative Agreement W911NF-10-2-0016.}

{\small
\bibliographystyle{ieee_fullname}
\bibliography{main}
}

\clearpage

\appendix
\appendixpage

Here we provide additional experimental results and details. We begin by evaluating the effect of the number of attributes used for training on the model's few-shot performance in Appendix~\ref{sec:attr_num}. Next, we explore the effect of plugging our compositional representations into existing few-shot learning methods, such as Prototypical Networks~\cite{snell2017prototypical} and Matching Networks~\cite{vinyals2016matching}, as well as the effect of data augmentation on these methods in Appendix~\ref{sec:exp_add}. We then provide a qualitative analysis of the learned representation using Network Dissection~\cite{zhou2018interpreting} in Appendix~\ref{sec:netdissect}. In Appendix~\ref{sec:depth} we demonstrate how to apply our proposed regularization to deeper network architectures. Finally, we visualize the attributes used in our experiments on ImageNet together with their hierarchical structure in Appendix~\ref{sec:attr}, and provide additional implementation details in Appendix~\ref{sec:impl}.% that are not included in the main paper due to limited space

\section{Effect of the Number of Attributes}
\label{sec:attr_num}
\begin{figure}[t!]
\centering
\includegraphics[width=.485\columnwidth]{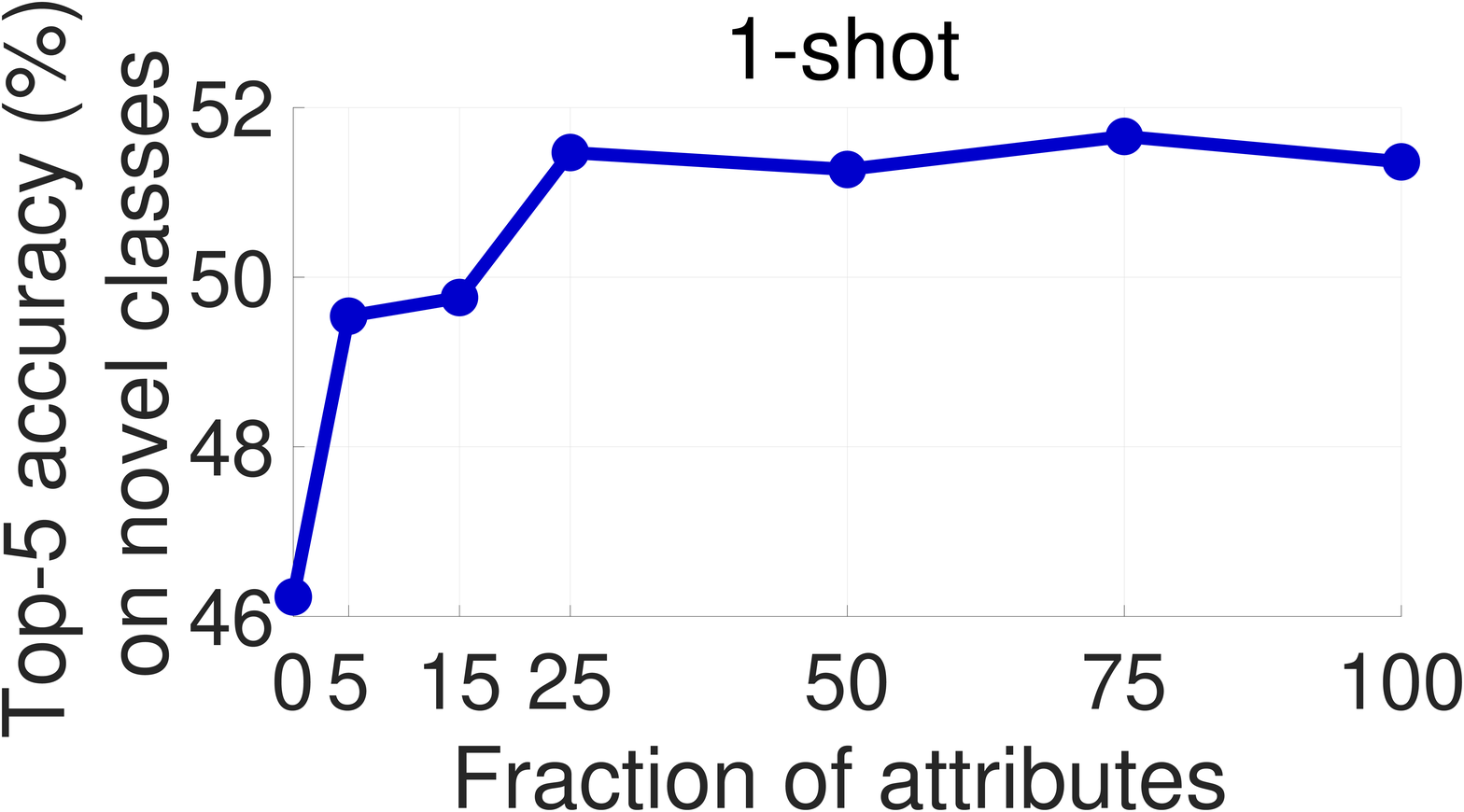}
~
\includegraphics[width=.485\columnwidth]{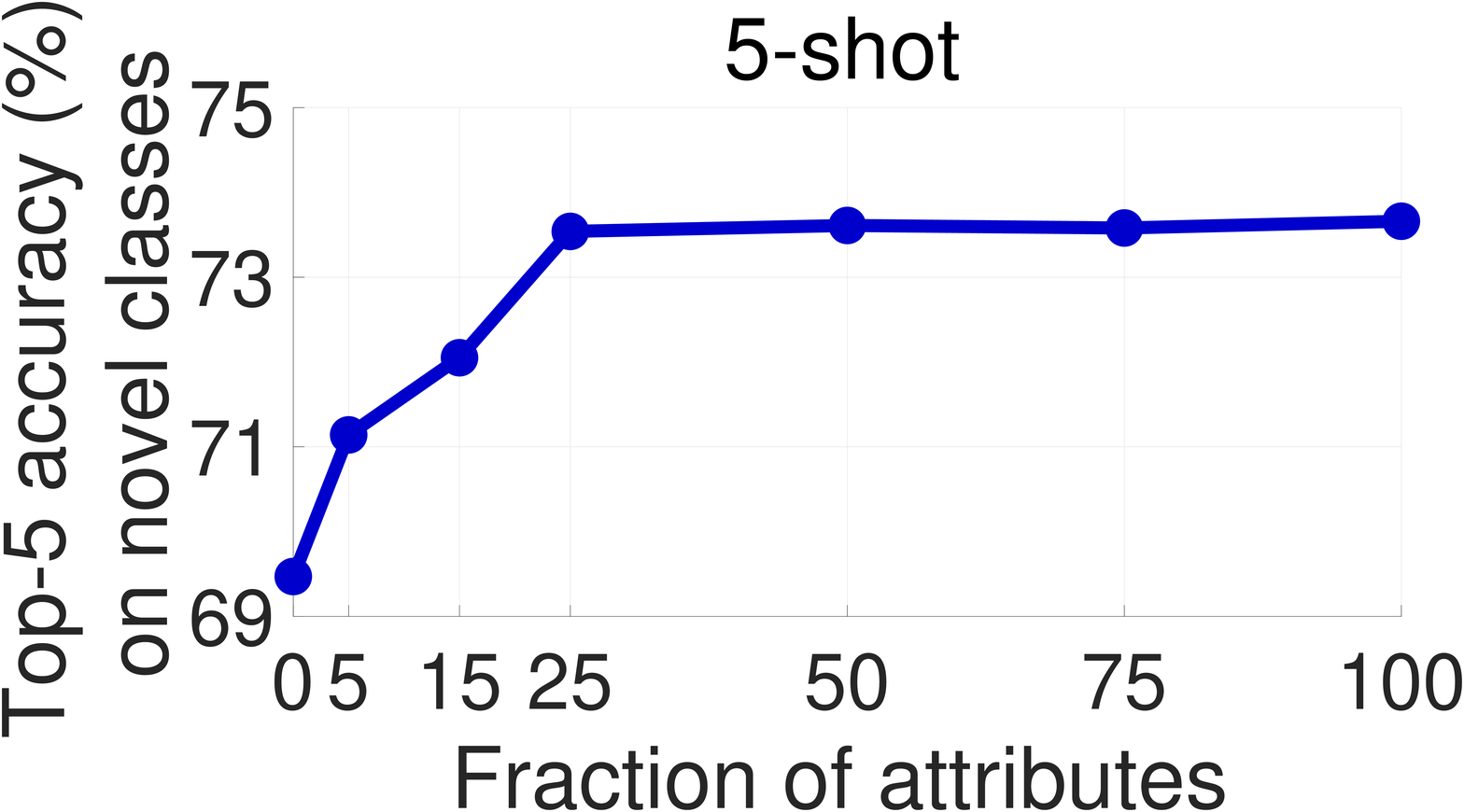}
   \caption{Evaluation of the effect of the number of attributes on the model's performance on the novel categories of the CUB dataset. Our approach provides significant improvements over the baseline even with as few as a quarter of the attributes.}
\label{fig:attrs_cub}
  \vspace{-4mm}
\end{figure}

\begin{figure}[t!]
\centering
\includegraphics[width=.485\columnwidth]{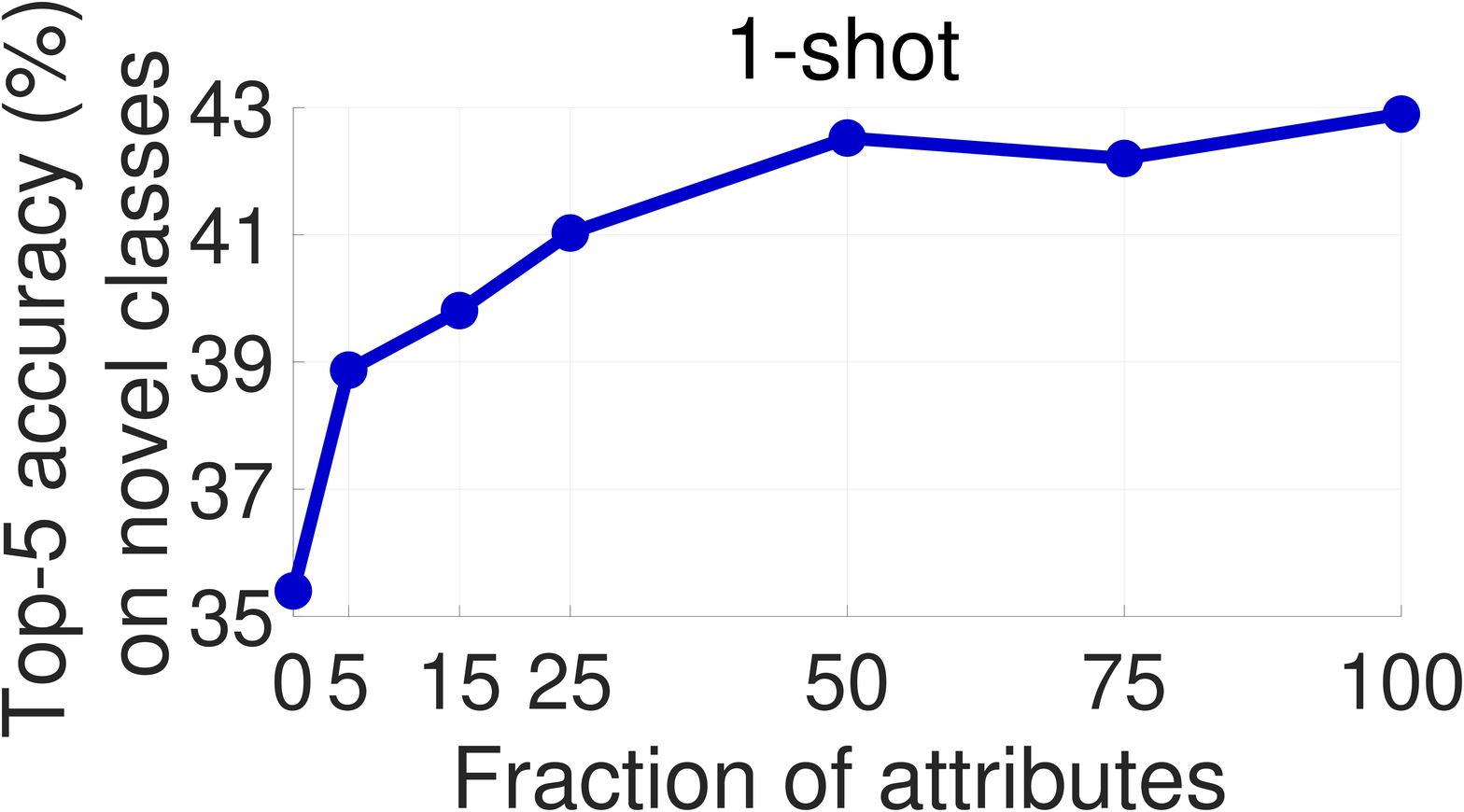}
~
\includegraphics[width=.485\columnwidth]{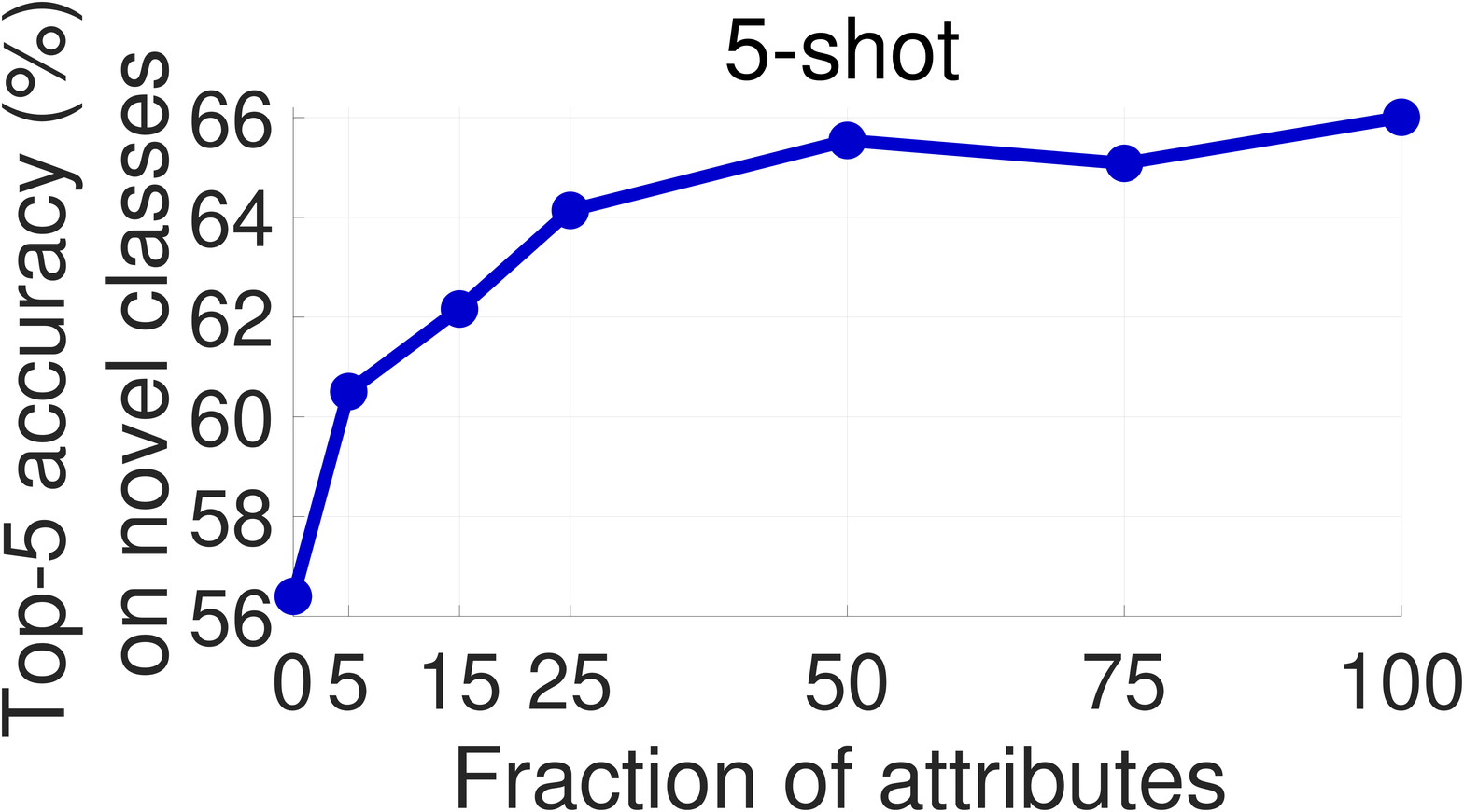}
   \caption{Evaluation of the effect of the number of attributes on the model's performance on the novel categories of the SUN dataset. In this case, half of the original number of attributes is enough to achieve comparable improvements over the baseline.}
  \vspace{-2mm}
\label{fig:attrs_sun}
\end{figure}

One of the main concerns with our proposed approach is that obtaining attribute supervision can be expensive in practice. To partially mitigate it, we study the effect of the number of attributes used in training on the model's few-shot learning performance. To this end, we sample 75/50/25/15/5\% of the attributes on CUB and SUN at random and train our models using these subsets with the soft constraint and orthogonality regularization. The results on the novel categories are presented in Figures~\ref{fig:attrs_cub} and~\ref{fig:attrs_sun}, respectively. Encouragingly, the performance decreases only slightly with the number of attributes. In particular, with only 25\% of the original attributes on CUB and 50\% on SUN (which correspond to 34 and 45 attributes, respectively), the performance hardly changes. Moreover, we achieve noticeable improvements over the 0-attribute baseline by using as few as 5 attributes on SUN. This result strengthens our claim that the proposed approach can improve the performance of few-shot learning methods with only a small annotation overhead. 

\section{Additional Analysis of Existing Few-Shot Learning Methods}
\label{sec:exp_add}
In the main paper, we have demonstrated that a simple  cosine classifier learned on top of a frozen CNN, which was trained with our compositionality regularization, leads to state-of-the-art results on three datasets, outperforming more complex existing few-shot classification models, such as Protoypical Networks~\cite{snell2017prototypical} and Matching Network~\cite{vinyals2016matching}. It is natural to ask whether training these models with our compositional representation would lead to superior results. To answer this question, we train the CNN backbone on the base categories with a linear classifier and the compositionality regularization. On top of the compositional feature, we learn these few-shot models as described in the main paper. We report the results on the novel categories of the CUB-200-2011 dataset in Table~\ref{tab:comp_few}.%Using features trained with a  cosine classifier resulted in few-shot learning methods failing to converge. 

\begin{table}[bt]
 \centering
  {%
    \footnotesize
\resizebox{.8\linewidth}{!}{
    \begin{tabular}{l|c@{\hspace{1em}}c@{\hspace{1em}}c@{\hspace{1em}}}
      & \multicolumn{3}{c}{Novel} \\
        & 1-shot          & 2-shot           & 5-shot      
              \\\hline
    PN  
        & 43.2       & 54.3       & 67.8     \\
    PN + data aug 
        & 44.0       & 54.8       & 68.1     \\
    PN w/ comp 
        & 41.5       & 55.0       & 68.4       \\
    PN w/ comp + data aug
        & 42.2       & 55.7       & 68.7       \\\hline
    MN
        & 48.5 & 57.3       & 69.2         \\
    MN + data aug
        & 49.3 & 57.9       & 69.7         \\
    MN w/ comp 
        & 50.9 &  59.5       & 71.6              \\
    MN w/ comp + data aug
        & 51.6 &  59.9       & 72.0             \\\hline
    Linear w/ comp  
        & 47.0      & 60.0       & 74.0       \\
     Cos w/ comp  
        & 52.5       & 63.6       & 73.8     \\
     Cos w/ comp + data aug 
        & \bf{53.6}       & \bf{64.8}       & \bf{74.6}     \\
\end{tabular}
}
}
\caption{Incorporating our compositional representations into existing few-shot classification models : top-5 accuracy on the novel categories of the CUB dataset. `PN': Prototypical Networks, `PN w/ comp': Prototypical Networks with our compositional representation, `MN': Matching Networks, `MN w/ comp': Matching Networks with our compositional representation, `Linear w/ comp': our compositional representation with a linear classifier, `Cos w/ comp': our compositional representation with a  cosine classifier. The variants trained with data augmentation are marked with `+ data aug'.}%learning methods the linear variant of 
\vspace{-2mm}
\label{tab:comp_few}
\end{table}

We observe that using compositional representations indeed leads to improved performance for both Prototypical Networks and Matching Networks in almost all the settings. The improvements for Prototypical Networks are marginal. The effect of compositional representations for Matching Networks is more pronounced, allowing them to outperform the linear classifier in the 1-shot evaluation setting. However, our  cosine classifier remains superior to the few-shot learning methods. In addition, compared with our approach, data augmentation has a less effect on the performance improvement of traditional few-shot learning methods. Our approach again consistently outperforms the baselines with data augmentation. These experiments not only confirm the surprising effectiveness of the  cosine classifier observed in the main paper, but also show that the proposed compositional representations can generalize to other scenarios and classification models.%the variant with a

\section{Qualitative Analysis of Compositional Representations}
\label{sec:netdissect}
We now qualitatively and quantitatively analyze the learned representations using Network Dissection: a framework for studying the interpretability of CNNs proposed by Zhou \etal~\cite{zhou2018interpreting}. They first collect a large dataset of images with pixel-level annotations, where the set of labels spans a diverse vocabulary of concepts from low-level (\eg, textures) to high-level (\eg, object and scene categories) concepts. They then probe each unit in a pre-trained CNN by treating it as a classifier for each of these concepts. If a unit achieves a score higher than a threshold for one of the concepts, it is assumed to capture the concept. The number of internal units that capture some interpretable concepts is then used as a measure of the interpretability of the network.

We compute this measure for the last layer of our networks (before the classification layer) for both the  cosine classifier baseline and the  cosine classifier with our compositionality regularization trained on SUN397. We observe that the baseline has 169 interpretable units out of 512, capturing 92 unique concepts. For our proposed compositional model, the number of interpretable units increases to 333 and the number of unique concepts increases to 119. Clearly, the proposed regularization results in learning a much more interpretable representation. To further analyze its properties, we present the distribution of the interpretable units for the baseline in Figure~\ref{fig:nds_base} and that for the proposed model in Figure~\ref{fig:nds_base_comp}, grouped by the concept type. We observe that our improvement in novel concepts mainly comes from the scene and object categories. %This is expected, since SUN397 is a scene classification dataset. 

\begin{landscape}
\begin{figure}[t]
\begin{center}
\includegraphics[width=1\linewidth]{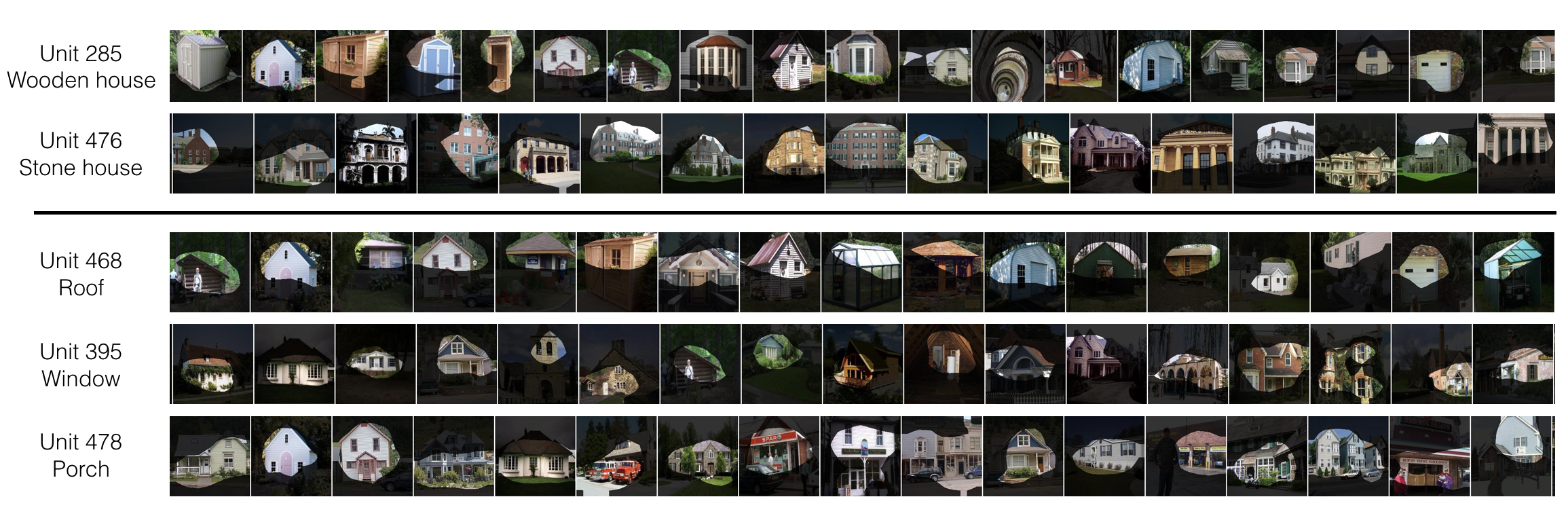} \vspace{0.1cm}
\end{center}
\vspace{-.6cm}
\caption{Top activating images for several units in the last layer of the network that are mapped to the concept {\tt house} by Network Dissection, together with the units' attention maps. The first two units are found both in the baseline model and in the model trained with our compositionality regularization, and capture generic concepts: {\tt wooden house} and {\tt stone house}. The next three units are only found in the proposed model and capture parts of the house, such as {\tt roof}, {\tt window}, and {\tt porch} (see attention maps).}
\label{fig:feat}
\vspace{-0.5cm}
\end{figure}
\end{landscape}

Another interesting observation is that most of the new interpretable units seem to be duplicates of the units that already existed in the baseline model. This is due to a limitation of the Network Dissection approach. Although the vocabulary of concepts which this evaluation can identify is relatively broad, it is still limited. Several different real-world concepts thus end up being mapped to a single label in the vocabulary. To illustrate this observation and further analyze our approach, we visualize the maximally activating images for several units that are mapped by Network Dissection to the category {\tt house} in Figure~\ref{fig:feat}.  The figure also shows attention maps of the units within each image. The first two units, which are shared by the baseline and the proposed model, seem to capture the general concepts of a {\tt wooden house} and a {\tt stone house}. However, the other three units, which are only found in the model trained with the compositionality regularization, seem to capture parts of the house, such as {\tt roof}, {\tt window}, and {\tt porch} (see attention maps). This observation further validates that our proposed approach leads to learning representations that capture the compositional structure of the concepts.

\section{Effect of the Network Depth} 
\label{sec:depth}
Up till now we used a ResNet-10 backbone for all the experiments. In this section, we study the generalizability of the proposed compositionallity regularization to deeper network architectures. We conduct these experiments on the SUN397 dataset due to its large size and high quality of the attribute annotations. In Table~\ref{tab:anal_deep} we compare the ResNet-10 model with  cosine classifier to ResNet-18 and ResNet-34. First of all, we notice that the improvements with respect to the baseline due to compositionallity regularization are diminishing as the network depth increases. Moreover, the shallow `ResNet-10, Cos w/ comp' model outperforms the deeper variants. We analyze this behavior and observe that the deeper models are able to learn attribute classifiers without significantly modifying their representation. This can be explained by the fact that the feature space of the last layer of the deep networks has a higher representation capability. We thus propose to adapt our regularization by applying it not only to the last, but also to the intermediate layers of the network. In practice, we apply it to the outputs of all the ResNet blocks starting from the block 9. This new variant, denoted as `Cos w/ deep comp'', achieves improvements over the baseline for ResNet-18 and ResNet-34, which is comparable to those of `Cos w/ comp' for ResNet-10. Such results confirm that our proposed approach is indeed applicable to deeper networks. The improvement is somewhat smaller for the novel classes though; all the three models perform approximately the same in this setting.%power
\begin{table}[bt]
 \centering
  {%
   % \footnotesize
\resizebox{\linewidth}{!}{
    \begin{tabular}{l|c@{\hspace{1em}}c@{\hspace{1em}}c@{\hspace{1em}}|c@{\hspace{1em}}c@{\hspace{1em}}c@{\hspace{1em}}c}
      & \multicolumn{3}{c|}{Novel} &
          \multicolumn{3}{c}{All}\\
        & 1-shot          & 2-shot           & 5-shot      
        & 1-shot           & 2-shot            & 5-shot         \\\hline
    ResNet-10, Cos
        & 35.4       & 45.6       & 56.4    
        & 52.1       & 56.7       & 61.9     \\
    ResNet-10, Cos w/ comp
        & 43.4       & 54.5       & \bf{65.9}      
        & 54.9       & 60.4       & 66.3       \\ \hline
    ResNet-18, Cos
        & 37.7       & 47.5       & 58.8    
        & 53.7       & 58.0       & 63.3     \\
    ResNet-18, Cos w/ comp
        & 41.2       & 51.6       & 63.0      
        & 55.5       & 60.6       & 66.3       \\
     ResNet-18, Cos w/ deep comp
        & \bf{43.9}       & \bf{54.7}       & 65.7      
        & 56.5       & 62.0       & 67.6       \\ \hline
    ResNet-34, Cos
        & 38.5       & 48.8       & 60.2    
        & 54.3       & 58.8       & 64.4     \\
    ResNet-34, Cos w/ comp
        & 41.0       & 51.5       & 62.5      
        & 56.1       & 60.9       & 66.3       \\
     ResNet-34, Cos w/ deep comp
        & 43.1       & 53.7       & 65.7      
        & \bf{57.0}       & \bf{62.1}       & \bf{68.2}      \\
\end{tabular}
}
}
\caption{Evaluation of deeper architectures: top-5 accuracy on the novel and all (\ie, $novel+base$) categories of the SUN dataset. `Cos': the baseline with a  cosine classifier, `Cos w/ comp': our proposed compositional representation with a  cosine classifier, `Cos w/ deep comp': our proposed compositional representation with regularization applied to intermediate layers of the network.}
\vspace{-2mm}
\label{tab:anal_deep}
\end{table}

\section{ImageNet Attributes}
\label{sec:attr}
\begin{landscape}
\begin{figure}[t]
\begin{center}
\includegraphics[clip,width=0.75\linewidth]{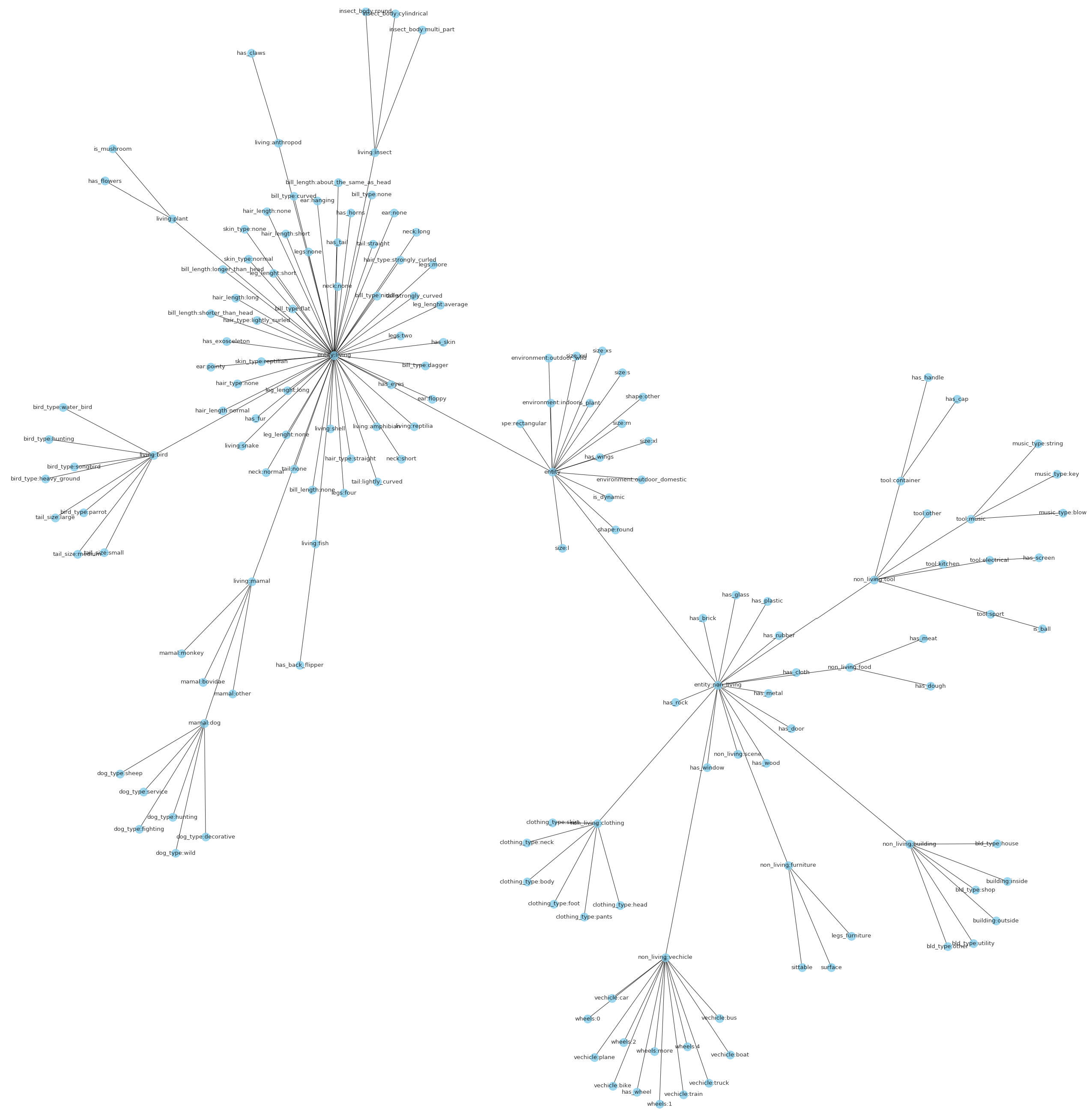} \vspace{0.1cm}
\end{center}
\vspace{-.3cm}
\caption{Attributes used in our ImageNet experiments together with their hierarchical structure. Each node represents a binary attribute and edges capture the parent-child relationships between the attributes.}
\label{fig:attrs}
\vspace{-0.5cm}
\end{figure}
\end{landscape}

In Figure~\ref{fig:attrs}, we visualize the hierarchical structure of the attributes which we defined for the 389 base categories in the subset of ImageNet used in our experiments. Each node (including non-leaf nodes) represents a binary attribute and edges capture the parent-child relationships between the attributes. These relationships are used in the annotation process to prune irrelevant attributes (such as number of wheels for a living thing) and thus save the annotator's time. Note that our annotated attributes might not be the perfect set of attributes for ImageNet. Nevertheless, even with these imperfect attributes, our compositionality regularization approach allowed us to achieve the state-of-the-art result.%We want to emphasize that the annotations have been collected with purely exploratory purposes, and we do not claim that this is the perfect set of attributes for ImageNet.

\section{Additional Implementation Details}
\label{sec:impl}
\paragraph{Training Schedules.} On ImageNet we use the setting proposed in~\cite{hariharan2017low,wang2018low}, with a batch size of 256 and 90 training epochs. The learning rate is decreased by a factor of 10 every 30 epochs. On SUN397 we use the same batch size and total number of epochs, but decrease the learning rate after the first 60 epochs, and then again after 15 more epochs. On CUB-200-2011, which is a much smaller dataset, we use a batch size of 16 and train for 170 epochs. The learning rate is first decreased by a factor of 10 after 130 epochs, and then again after 20 more epochs. These schedules are selected on the validation set.%Hyper-Parameters 

\paragraph{Compositionality Regularization.} On ImageNet the trade-off hyper-parameter of the compositionality regularizer $\lambda$ is set to 8, on SUN397 to 25, and on CUB-200-2011 to 15. The hyper-parameter of the orthogonality constraint $\beta$ is set to 0.001, 0.0025, and 0.00035 on the three datasets, respectively. To select these values, we split the base categories in half, using one half as validation. The attribute annotations are sparse, with around 10\% of them being labeled as positive for any given image on average.  Due to this highly imbalanced distribution of training labels, all the attribute classifiers learn to predict the negative labels. To address it, we randomly sample a subset of the negative attributes for every example in every batch to balance the number of positive and negative examples.%weight each

\paragraph{Few-Shot Training.} On ImageNet we train for 100 iterations for both  cosine and linear classifiers. On SUN397 we train for 200 iterations for linear and 100 for  cosine classifier. On CUB-200-2011 we train for 100 iterations for linear and 40 for  cosine classifier. To select these values, we use the same split of the base categories discussed above, and train until the top-5 performance on the validation categories stops increasing. We use the same setting to select the optimal hyper-parameters for other methods. 

\begin{landscape}
\begin{figure}[t]
\begin{center}
\includegraphics[clip,width=1\linewidth]{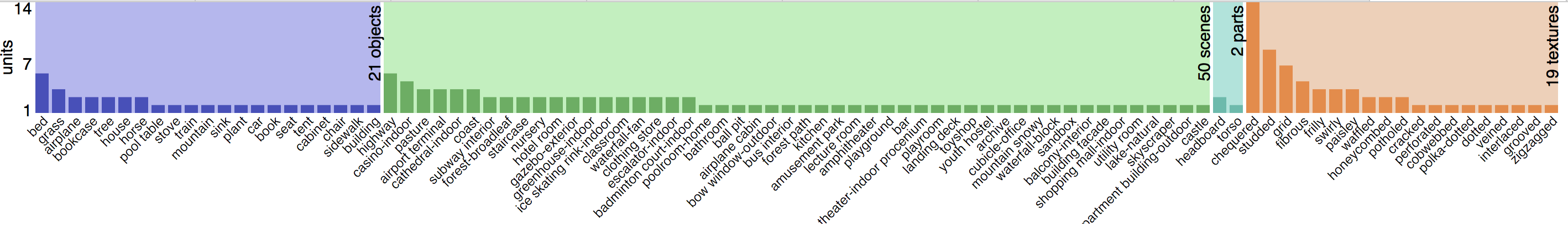} \vspace{0.1cm}
\end{center}
\caption{Distributions of interpretable units in the last layer of the baseline model trained with a  cosine classifier on SUN397 according to Network Dissection~\cite{zhou2018interpreting}. The units are grouped by the type of the concepts they represent (\ie, object, scene, part, or texture). Overall, this layer has \textbf{169} interpretable units, capturing \textbf{92} unique concepts.}
\label{fig:nds_base}
\vspace{-0.5cm}
\end{figure}

\begin{figure}[t]
\begin{center}
\includegraphics[clip,width=1\linewidth]{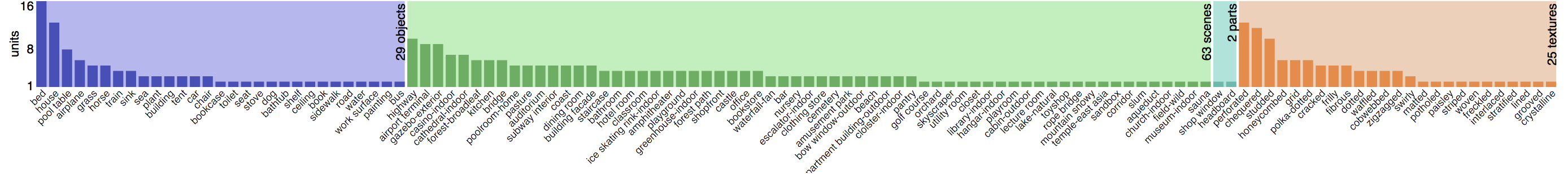} \vspace{0.1cm}
\end{center}
\caption{Distributions of interpretable units in the last layer of the model trained with a  cosine classifier and our compositionality regularization on SUN397 according to Network Dissection~\cite{zhou2018interpreting}. The units are grouped by the type of the concepts they represent (\ie, object, scene, part, or texture). Overall, this layer has \textbf{333} interpretable units, capturing \textbf{119} unique concepts.}
\label{fig:nds_base_comp}
\vspace{-0.5cm}
\end{figure}
\end{landscape}

\end{document}